\documentclass[review]{elsarticle}

\usepackage{amsmath,graphicx,amsfonts,caption} %
\usepackage{amssymb}
\usepackage[hidelinks,bookmarks=false]{hyperref}
\usepackage{xcolor}

\journal{Signal Processing: Image Communication}

\DeclareMathOperator{\sinc}{sinc}

\begin{document}

\begin{frontmatter}

\title{Spectral analysis of re-parameterized light fields}

\author{Martin~Alain}
\corref{mycorrespondingauthor}
\author{Aljosa~Smolic}
\address{V-SENSE\fnref{myfootnote}, Trinity College Dublin}
\fntext[myfootnote]{This publication has emanated from research conducted with the financial support of Science Foundation Ireland (SFI) under the Grant Number 15/RP/2776.}

\begin{abstract}
In this paper, we study the spectral properties of re-parameterized light field.
Following previous studies of the light field spectrum, which notably provided sampling guidelines, we focus on the two plane parameterization of the light field.
However, we introduce additional flexibility by allowing the image plane to be tilted and not only parallel.
A formal theoretical analysis is first presented, which shows that more flexible sampling guidelines (i.e. wider camera baselines) can be used to sample the light field when adapting the image plane orientation to the scene geometry.
We then present our simulations and results to support these theoretical findings.
While the work introduced in this paper is mostly theoretical, we believe these new findings open exciting avenues for more practical application of light fields, such as view synthesis or compact representation.
\end{abstract}


\end{frontmatter}


\section{Introduction}

\begin{figure}[t]
	\centering
	\includegraphics[width=\linewidth,trim={0cm 4cm 1.5cm 0cm},clip]{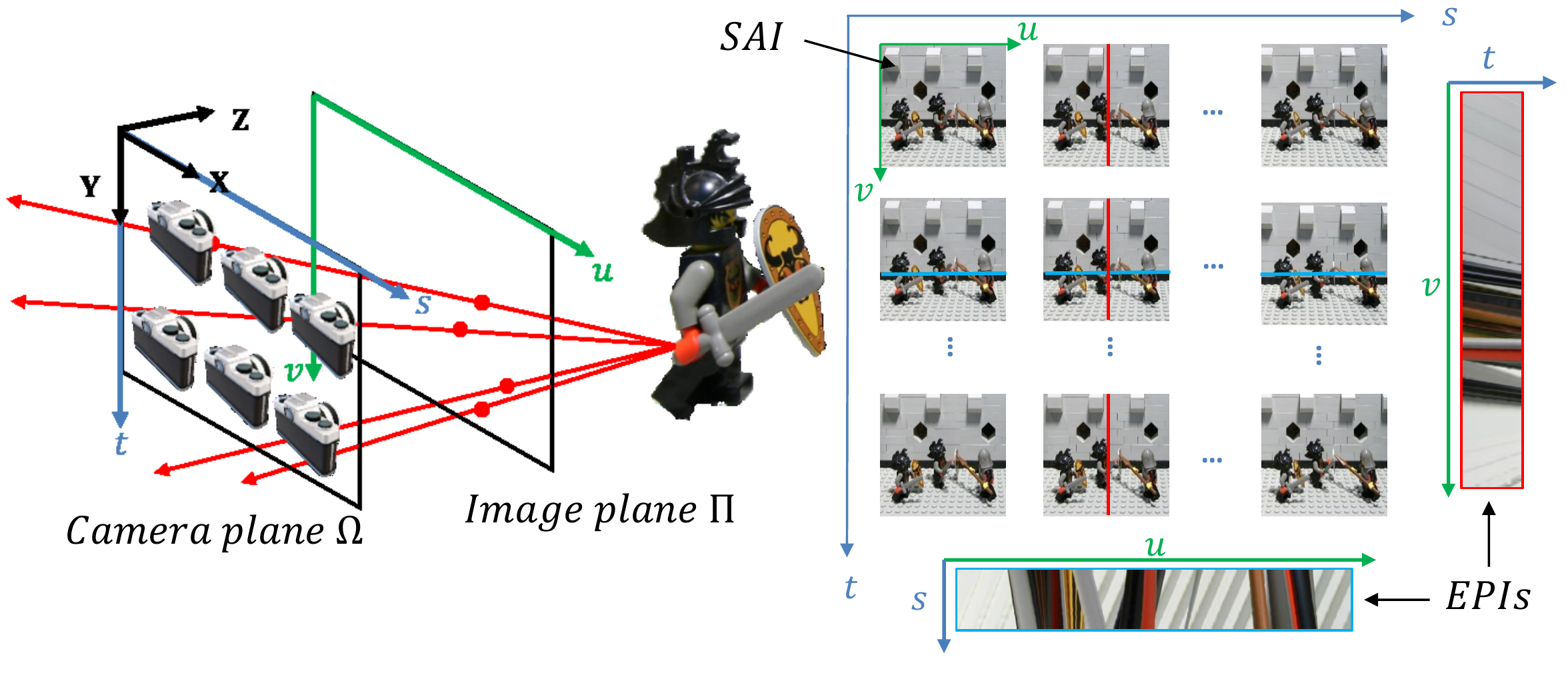}

	\caption{Two-parallel plane parameterization of the light field.
	The light field can be represented as a matrix of sub-aperture images (SAI); or Epipolar Plane Images (EPI) shown below and on the right.}

	\label{fig:light_field_geo}

\end{figure}

The study of the spectral properties of the plenoptic function has been ongoing for at least two decades and was proven useful for both theoretical understanding as well as practical aspects of image-based rendering~\cite{chai2000plenoptic}.
The concept of the plenoptic function was introduced by Adelson and Bergen in  \cite{Adelson91} as a complete representation of light rays in space-time, and can be formally described as a 7 dimensional function $(x, y, z, \alpha, \beta, \lambda, \tau) \to p(x, y, z, \alpha, \beta, \lambda, \tau)$, measuring the intensity of light for every point in space $(x, y, z)$, for every possible angle $(\alpha, \beta)$, for every wavelength $\lambda$, at any time $\tau$.
While an important theoretical tool, practical sampling of the complete plenoptic function remains a difficult challenge.
As an alternative, the concept of light fields has been introduced to reduce the dimensionality of the plenoptic function, based on simpler assumptions which are usually not limiting in practice, such as considering that the intensity of light rays remain constant along a straight line.
The 4D two-parallel plane light field parameterization was notably introduced simultaneously around 25 years ago by Levoy and Hanrahan \cite{Levoy1996}, and Gortler et al. \cite{gortler1996lumigraph}.
This is the main parameterization used in previous studies on the light field spectrum~\cite{chai2000plenoptic,zhang2003spectral,do2011bandwidth,gilliam2013spectrum}, and the one we also adopt in this paper.
As shown in Fig.~\ref{fig:light_field_geo}, we denote in this paper the 4D light field as a 4D function $\Omega \times \Pi \to \mathbb{R}, (s, t, u, v) \to p(s, t, u, v)$ in which the plane $\Pi$ represents the spatial distribution of light rays indexed by $(u, v)$, also called the image of focal plane, while $\Omega$ corresponds to their angular distribution indexed by $(s, t)$, also called the camera plane.
The light field can be visualized as a regular grid of viewpoint images, or views (see Fig. \ref{fig:light_field_geo}). 
Each view represents a 2D slice of the light field over the spatial dimensions ($u, v$).
Another common representation of light fields are Epipolar Plane Images (EPI), which are 2D slices of the 4D light field obtained by fixing one spatial and one angular dimension ($su$- or $vt$-planes, see Fig. \ref{fig:light_field_geo}), and have been widely used to study the spectral properties of light fields.

\begin{figure}[t]
	\centering
	\includegraphics[width=0.9\linewidth,trim={0cm 1.1cm 0cm 3cm},clip]{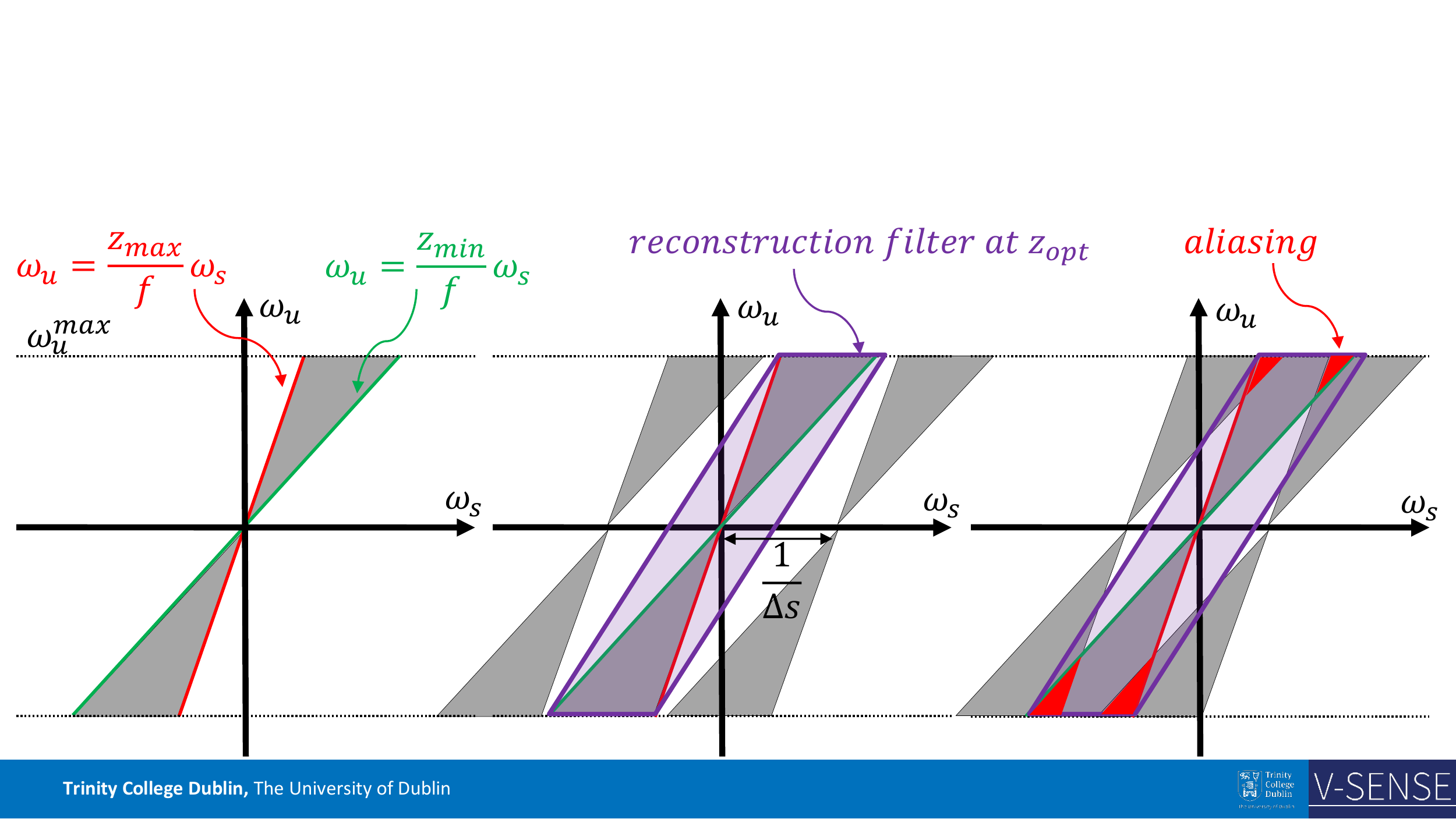}

	\caption{Left: The 2D light field spectrum support is fan-shaped and the boudaries depend on the scene depth range $[z_{min},z_{max}]$ and the ligh field cameras focal length $f$. Center: With a suitable reconstruction filter and camera spacing $\Delta s \leq \Delta s_{max}$, the light field can be correctly sampled. Right: If the camera spacing is too large $\Delta s > \Delta s_{max}$, aliasing occurs.}
	\label{fig:EPISpectrumSampling}

\end{figure}

The light field spectral analysis of Chai et al. first showed that the light field spectrum in the Fourier domain is band limited, and its support is a ``bow-tie'' or fan-shaped~\cite{chai2000plenoptic}, with slopes proportional to the depth range of the scene (see Fig~\ref{fig:EPISpectrumSampling}).
Based on the Nyquist sampling theorem, Chai et al. use this analysis to provide light field sampling guidelines (hence the concepts of light field spectral analysis and plenoptic sampling are sometimes used interchangeably). In particular, the maximum bound for spacing the light field cameras $\Delta s_{max}$ (also called camera baseline) can be derived such that there is no overlap in-between the light field spectrum replicas caused by the sampling process, as shown in Fig~\ref{fig:EPISpectrumSampling}, and can be expressed as follows:

\begin{equation}
\label{eq:smax}
    \Delta s_{max} = \frac{1}{f (z_{min}^{-1} - z_{max}^{-1}) \omega_u^{max}}
\end{equation}

\noindent where $f$ is the focal length of the light field cameras, $z_{min}$ and $z_{max}$ correspond to the minimum and maximum bounds of the scene depth range respectively, and $\omega_u^{max}$ is the highest spatial frequency.
Note that we can consider that there exist a bound on the spatial frequency $\omega_u$ due to the finite pixel resolution and low-pass filtering of the camera optics.
In addition, the optimal depth $z_{opt}$ for a reconstruction filter with constant depth depending on the scene depth range is derived as:

\begin{equation}
\label{eq:zopt}
    z_{opt} = \frac{2}{z_{min}^{-1} + z_{max}^{-1}}
\end{equation}

\begin{figure}[t]
	\centering
	\includegraphics[width=0.9\linewidth,trim={0cm 1.1cm 0cm 2.8cm},clip]{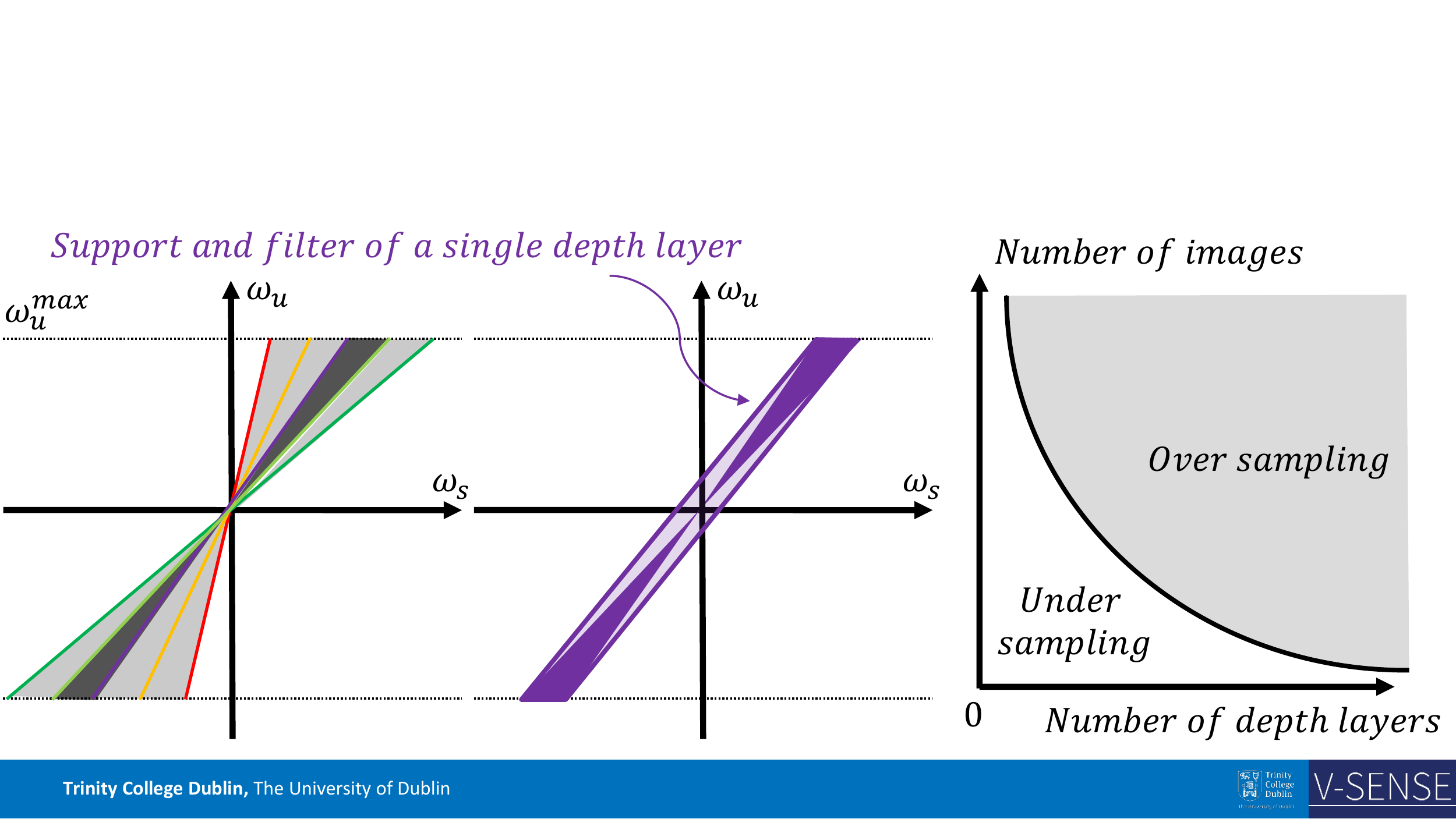} 
	\caption{Right: The minimum sampling curve introduced in \cite{chai2000plenoptic} shows that the number of light field images can be decreased if the knowledge of the scene geometry in the form of depth layers increases. Left and center: Each depth layer has a smaller depth range which results in reduced light field spectrum support which can be sampled with a larger camera spacing.}
	\label{fig:minimum_sampling_curve}

\end{figure}

Note that this analysis is carried in the 2D $su$-EPI domain, but can be generalised to the full 4D light field.
Similar results were derived through a geometric analysis of light field rendering in~\cite{lin2004geometric}.
In addition, Chai et al. study in~\cite{chai2000plenoptic} the minimum sampling curve in the presence of a known geometric proxy, namely depth layers.
By dividing the scene into non-overlapping depth layers which cover a fraction of the full depth range, the Fourier spectrum support of each layer is also reduced to a fraction of the full light field spectrum, as shown in Fig.~\ref{fig:minimum_sampling_curve}.
This allows to increase the camera baseline for capturing the light field and reduces the total number of views.
This concept of joint sampling in the image and geometry space is summarized in the now well-known sampling curve shown in Fig.~\ref{fig:minimum_sampling_curve}.

The light field spectral analysis was explicitly extended to the full 4D light field by Dansereau et al., which showed that the generalised 4D spectrum support shape is a hyperfan~\cite{dansereau2015linear}.
By designing hyperfan shaped filters, light field denoising or volumetric refocusing can be performed.

While the work described above pioneered the study of plenoptic sampling, it is based on assumptions that the scene is Lambertian and free of occlusions, which is often not the case in practice.
Following work by Zhang and Chen showed that the light field spectrum support expands when considering non-Lambertian surfaces and occluders~\cite{zhang2003spectral}.

In such cases, it is assumed that light rays emitted from a same point change slowly with respect with their angular direction, so that it can be approximated with a band-limited signal.
The fan-shaped spectrum of the light field is then expanded by the bandwith of the non-Lambertian signal.
For scene with unknown geometry and occlusions, the scene is modeled as a set of planar objects parallel to the light field plane, i.e. each object has a constant depth.
It is shown that the spectrum of occluded objects is modulated by the spectrum of the occluding object.
A finer analysis of occlusions was recently proposed by Zhu et al. based on an occlusion field model~\cite{Zhu2020}.

However, it was later demonstrated by Do et al. that the light field spectrum is, in general, \emph{not} bandwidth-limited~\cite{do2011bandwidth}.
Thus, they introduce the concept of essential-bandwidth to study a wider range of scenes relying on less limiting assumptions.
In particular, while previous studies mostly modeled scenes with planar objects, they demonstrate that the slope of the objects provide a more accurate estimate of the light field bandwidth.
For this purpose they model the scene as a smooth surface painted with a band-limited signal (texture).
In addition, they demonstrate that by shearing the light field EPI, it is possible to obtain a spectrum most compacted near the spatial frequency axis.
Shearing the EPI correspond to modifying the depth of the light field image plane, and the optimal shearing correspond to placing the image plane at the depth $z_{opt}$ as defined in equation~\ref{eq:zopt}.

Following Do et al., Gilliam et al. used the concept of essential-bandwidth to carry an in-depth study of scenes containing a tilted plane of finite length painted with a band limited signal~\cite{gilliam2013spectrum}.
Using this scene model, they derive a closed-form expression of the light field spectrum.
They demonstrate in particular that the optimal depth $z_G$ of the reconstruction filter for a scene containing a single tilted plane is different than in the general case $z_{opt}$, and can be expressed as:

\begin{equation}
\label{eq:zG}
    z_{G} = \frac{z_{min} + z_{max}}{2}
\end{equation}

\noindent The geometry of more complex scene is then modeled as a set of tilted planes, and the spectrum of the scene can then be computed due to the linearity of the Fourier transform.
It should be noted however that this model is limited to continuous surfaces without occlusions.

Beyond sampling guidelines, the analysis of the light field spectrum has inspired a novel light field representation based on the shearlet transform~\cite{vagharshakyan2017light}.
The shearlet transform allows to tile the frequency plane with trapezoid-shaped tiles which are especially adapted to fit the sheared fan-shaped support of the light field spectrum.
The sparsity of the light field in the shearlet domain is used by Vagharshakyan et al. as a regularization constraint to reconstruct dense light fields from sparsely-sampled light fields.

More recently, Le Pendu et al. introduced the Fourier Disparity Layer (FDL) representation for light fields~\cite{le2019fourier}, which models the scene as a set of layers at constant depth, similar in principle to the concept introduced by Chai et al.~\cite{chai2000plenoptic} described above.
However, the scene depth does not need to be explicitly known, instead
the properties of this model in the Fourier domain are used to solve the inverse problem of estimating the FDL representation from an input light field which is assumed to be densely sampled and free of occlusions.
The FDL can then be used to efficiently render views with novel viewpoints and/or synthetic depth-of-field effects (i.e. refocusing).

Another recent layered representation is the Multiple Plane Images (MPI) developed for high quality real-time view synthesis from light fields~\cite{srinivasan2019pushing,mildenhall2019local,broxton2020immersive}, which also finds theoretical grounds in the minimum sampling curve of Chai et al. (see~Fig.\ref{fig:minimum_sampling_curve}), allowing to reduce the number of light field views if multiple depth layers are known, which was verified experimentally~\cite{mildenhall2019local}.
While the original MPI representation is based on regularly spaced planes, recent work explores adaptive placement of the MPI for a more efficient and compact representation~\cite{Navarro2021}.

In this paper, we propose to study the spectrum of the light field depending on its image plane parameterization, namely its position and orientation.
Note that changing the image plane position or orientation does not require a specific capture device or capturing multiple light field images, but is in practice achieved by transforming the individual light field views, which correspond to the re-parameterization of a single two-parallel plane light field.
This process has also been referred to as a change of basis in~\cite{Ng2005b}.
We first revisit the two-parallel plane light field spectral analysis, by explicitly taking into account the distance $D$ between the camera plane and a global images plane, which then determines the position of the local image planes usually considered when analysing the light field spectrum.
The derivation of the light field spectrum can then be carried out and studied depending on the distance $D$, and in particular we show it provides a new way of demonstrating results first introduced in~\cite{do2011bandwidth} on the optimal parameterization of the image plane to obtain the most compact spectrum.
We then extend this analysis to include the orientation of the global image plane, effectively removing the parallel plane constraint.
We show that by adapting the image plane orientation to the scene geometry, we can further reduce the size of the light field spectrum support, which can be useful in practice as it allows to sample the light field with larger camera baseline.

This paper is organized as follows. 
In section~\ref{sec:TPP}, we introduce the problem formulation, and the re-parameterization of the two-parallel plane light field is discussed. 
In section~\ref{sec:TIP}, we further extend the re-parameterization capabilities by allowing to tilt the image plane.
Experiments and results are presented in section~\ref{sec:results}.
Finally, we conclude and discuss future work in section~\ref{sec:conclusion}.

\section{Spectral analysis of the two-parallel plane light field parameterization}
\label{sec:TPP} 

\begin{figure}[t]
	\centering
	\includegraphics[width=\linewidth,trim={0cm 1.1cm 0cm 0cm},clip]{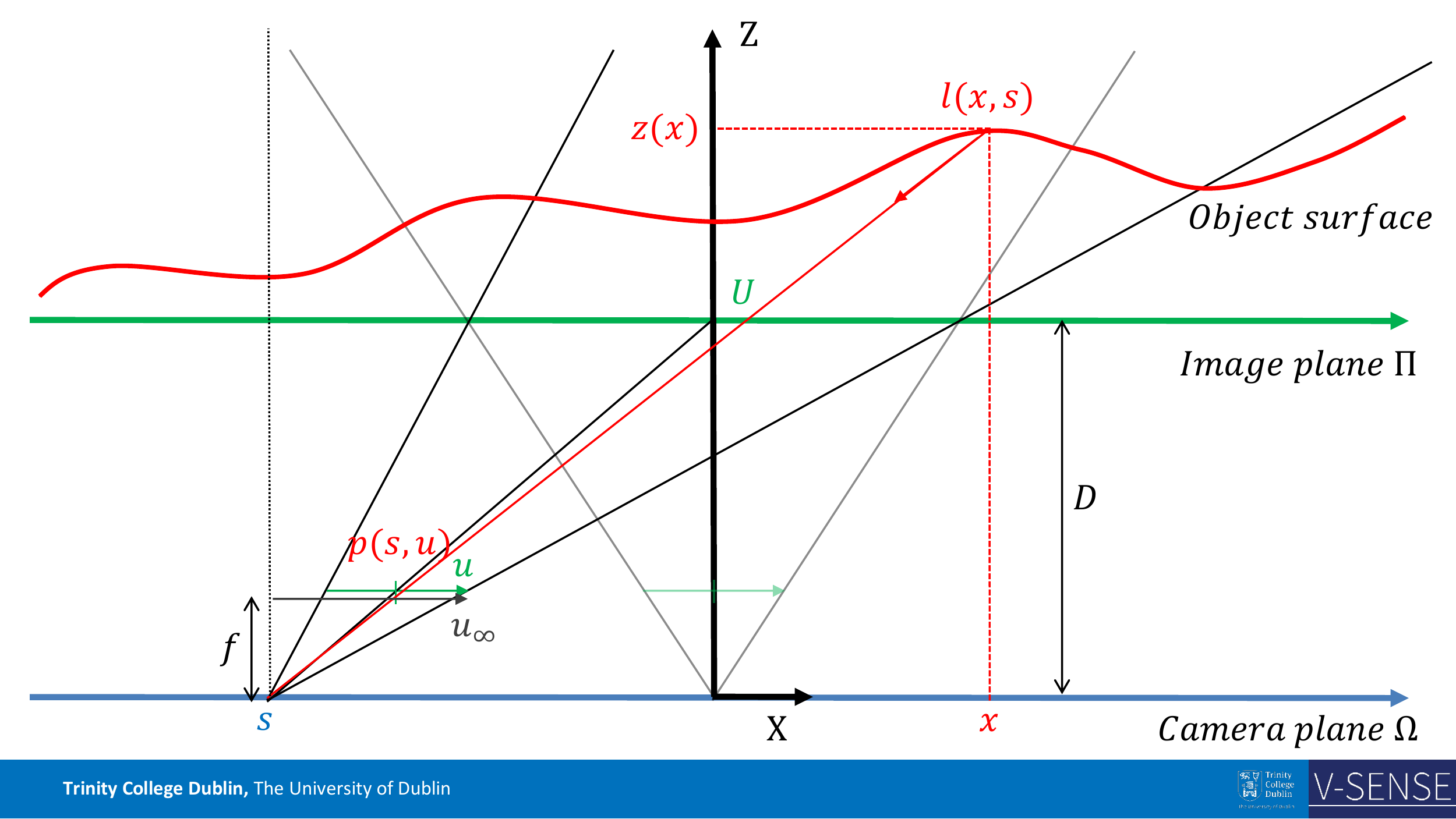}
	\caption{Two-parallel plane parameterization for a 2D slice of the light field using a global parameterization of the image plane $\Pi$.}
	\label{fig:TPP}
\end{figure}

\subsection{Scene model and parameterization of the image plane}

We first present the more detailed two-plane parameterization used in this paper, where a distinction is made between the local and global parameterization of the image plane.
For simplicity of notations, as in previous work~\cite{chai2000plenoptic,zhang2003spectral,do2011bandwidth,gilliam2013spectrum}, we carry our analysis for a 2D slice of the light field, i.e. the EPI over $s$ and $u$, as shown in Fig.~\ref{fig:TPP}.
Without loss of generality, we consider that the $x$ and $s$ axis are aligned and share a common origin.

The scene geometry is modelled as a surface parameterized with depth $z(x) > 0$, and the scene texture by a surface light field $l(x,s)$.
The angular direction of a light ray of $l$ emitted from the surface point $(x, z(x))$ is indexed with $s$, still without loss of generality.

The captured light field $p$ is indexed by $s$ on the camera plane and $u$ in the local image plane, and by construction we have $p(s,u)=l(x,s)$.
Our later goal is to analyse the light field spectrum $P(\omega_s,\omega_u)$ depending on the scene geometry and texture.

The two-parallel plane light field is parameterized such that the local image plane is positioned at depth $f$, the focal length.
Furthermore, we define the light field global image plane positioned at depth $D$, which is by definition where the light field camera frustums coincide (see Fig.~\ref{fig:TPP}).
Another way of understanding the parameterization used in this paper is to observe that, contrary to previous papers, the light field pinhole camera principal points are \emph{not} located at the centre of the local image plane, as shown in Fig.~\ref{fig:TPP}. (See also Fig.~\ref{fig:uinf_similar_triangle} for an example where all camera principal points are centered).
We denote the local image coordinate when all principal points are centred, i.e. when the global image plane is at infinity, by $u_{\infty}$.
We can then express $u_{\infty}$ depending on the generic local image coordinate $u$ as:

\begin{equation}
    \label{eq:u_inf}
    u_{\infty} = u - \frac{sf}{D}
\end{equation}

\noindent As observed in previous work, the geometry of the light rays as shown in Fig.~\ref{fig:TPP} can be expressed as:

\begin{equation}
    \label{eq:lf_geo_u_inf}
    z(x)u_{\infty} - xf + sf = 0
\end{equation}

\noindent By substituting the expression of $u_{\infty}$ from equation~\ref{eq:u_inf} (see appendix~\ref{sec:TPP_app} for more details), we obtain the generic geometric mapping linking a light ray $(s,u)$ to its point of origin on the object surface at position $(x, z(x))$:

\begin{equation}
    \label{eq:lf_geo}
        u = \frac{xf}{z(x)} + s f \left(\frac{1}{D} - \frac{1}{z(x)}\right)
\end{equation}

\noindent It is easy to verify that when $D \to \infty$, equation~\ref{eq:lf_geo} is reduced to equation~\ref{eq:lf_geo_u_inf} which is the geometric mapping used in previous work.
We assume that both $D > 0$ and $z(x) > 0$, thus the geometric mapping of equation~\ref{eq:lf_geo} is defined for all $x$.
In addition, we follow the no self-occlusion condition, which ensures a one-to-one mapping by constraining the object surface such that:

\begin{equation}
    \label{eq:no_self_occ}
    \lvert z'(x) \rvert < f \left(u_{max} + \frac{s_{max}f}{D}\right)^{-1}
\end{equation}

\noindent where $z'(x)$ is the derivative of $z$ with respect to $x$, the local image plane is limited by $\lvert u \rvert < u_{max}$, and the camera plane is limited by $\lvert s \rvert < s_{max}$. 
The expression in equation~\ref{eq:no_self_occ} is again different from the one used in \cite{do2011bandwidth,gilliam2013spectrum} as it explicitly involves the image parameter $D$. If $D \to \infty$, it reduces to the original formulation.

\subsection{Spectral analysis}

The one-to-one mapping provided by the geometric mapping of equation~\ref{eq:lf_geo} under the no self-occlusion condition of equation~\ref{eq:no_self_occ} then allows to connect the surface light $l$ to the captured light field $p$:

\begin{equation}
p(s,u) = l(x,s), \text{ with } u = \frac{xf}{z(x)} + s f \left(\frac{1}{D} - \frac{1}{z(x)}\right)
\end{equation}

Following~\cite{do2011bandwidth} and \cite{gilliam2013spectrum}, we can therefore  analyse the properties of light field spectrum $P$ depending on the properties of the surface light field spectrum $L$.
Starting from the definition of $P$, and using the change of variable based on equation~\ref{eq:lf_geo} (for detailed derivations, see appendix~\ref{sec:TPP_app}), we obtain the following expression:

\begin{equation}
\label{eq:P_fourier_tr}    
    P(\omega_s, \omega_u) =
    \int_{-\infty}^{\infty}  \frac{f}{z(x)^2} e^{-j\left(\omega_u \frac{xf}{z(x)}\right)} \left(H(x, \omega_s) * G(x, \omega_s)\right) dx
\end{equation}

\noindent where we define:

\begin{equation*}
    h(x,s) \triangleq (z(x) + z'(x) (s-x)) l(x,s)    
\end{equation*}
\begin{equation*}
    g(x,s) \triangleq e^{-j s \omega_u f (\frac{1}{D} - \frac{1}{z(x)})})    
\end{equation*}

\noindent with $H(x, \omega_s)$ and $G(x, \omega_s)$ the Fourier transform of $h$ and $g$ over $s$ respectively, and $*$ refers to the convolution over $\omega_s$.
Using the Fourier transform and convolution properties, we have (see appendix~\ref{sec:TPP_app} for more details):

\begin{equation}
\label{eq:HG}    
    H(x, \omega_s) * G(x, \omega_s)
    = H\left(x, \omega_s + \omega_u f (\frac{1}{D} - \frac{1}{z(x)})\right)
\end{equation}

By denoting $L(x, \omega_s)$ the Fourier transform of the surface light field $l$ over $s$, and using Fourier transform properties, we get the following expression:

\begin{equation}
\label{eq:H_fourier_tr}
\begin{split}
    H(x, \omega_s) = 
    (z(x) - x z'(x)) L(x, \omega_s) + j z'(x) \frac{\partial L(x, \omega_s)}{\partial\omega_s}
\end{split}
\end{equation}

We can now analyse the properties of $P(\omega_s, \omega_u)$ based on the properties of $L(x, \omega_s)$.
A common and reasonable assumption, which we adopt here, consists in considering that the surface light field $l$ emitted from a fixed surface point $(x, z(x))$ varies very slowly with respect to the angular direction $s$.
The surface light field is even often assumed to be Lambertian, in which case it does not change at all with $s$.
In terms of spectral properties, we can thus assume that $l(x,s)$ is bandwidth-limited with respect to $s$, which we can formally write as:

\begin{equation}
\label{eq:L_band_limit}
    L(x,\omega_s) = 0, \text{ if } \lvert \omega_s \rvert > B_L
\end{equation}

\noindent Note that in the case of a Lambertian light field, we have $B_L=0$. The no self-occlusion condition of equation~\ref{eq:no_self_occ} ensures that $ (z(x) - x z'(x)) > 0$, and from equations~\ref{eq:P_fourier_tr}, \ref{eq:HG}, \ref{eq:H_fourier_tr},  and \ref{eq:L_band_limit}, it follows that:

\begin{equation}
\label{eq:P_band_limit}
    P(\omega_s, \omega_u) = 0, \text{ if } \lvert \omega_s + \omega_u f (\frac{1}{D} - \frac{1}{z(x)}) \rvert > B_L
\end{equation}

Note that while the convolution of $H$ and $G$ has an analytical expression as shown in equation~\ref{eq:HG}, a more generic way of interpreting this operation is to consider that the bandwidth of the convolution of $H$ and $G$ is the sum of their respective bandwidth.
From equations~\ref{eq:H_fourier_tr} and \ref{eq:L_band_limit}, we get that an upper bound of the bandwidth of $H$ is $B_L$, while the bandwidth of $G$ is $\omega_u f (\frac{1}{D} - \frac{1}{z(x)})$, which also leads to the result of equation~\ref{eq:P_band_limit}.

\begin{figure}[t]
	\centering
	\includegraphics[width=0.9\linewidth,trim={0cm 1.1cm 0cm 3cm},clip]{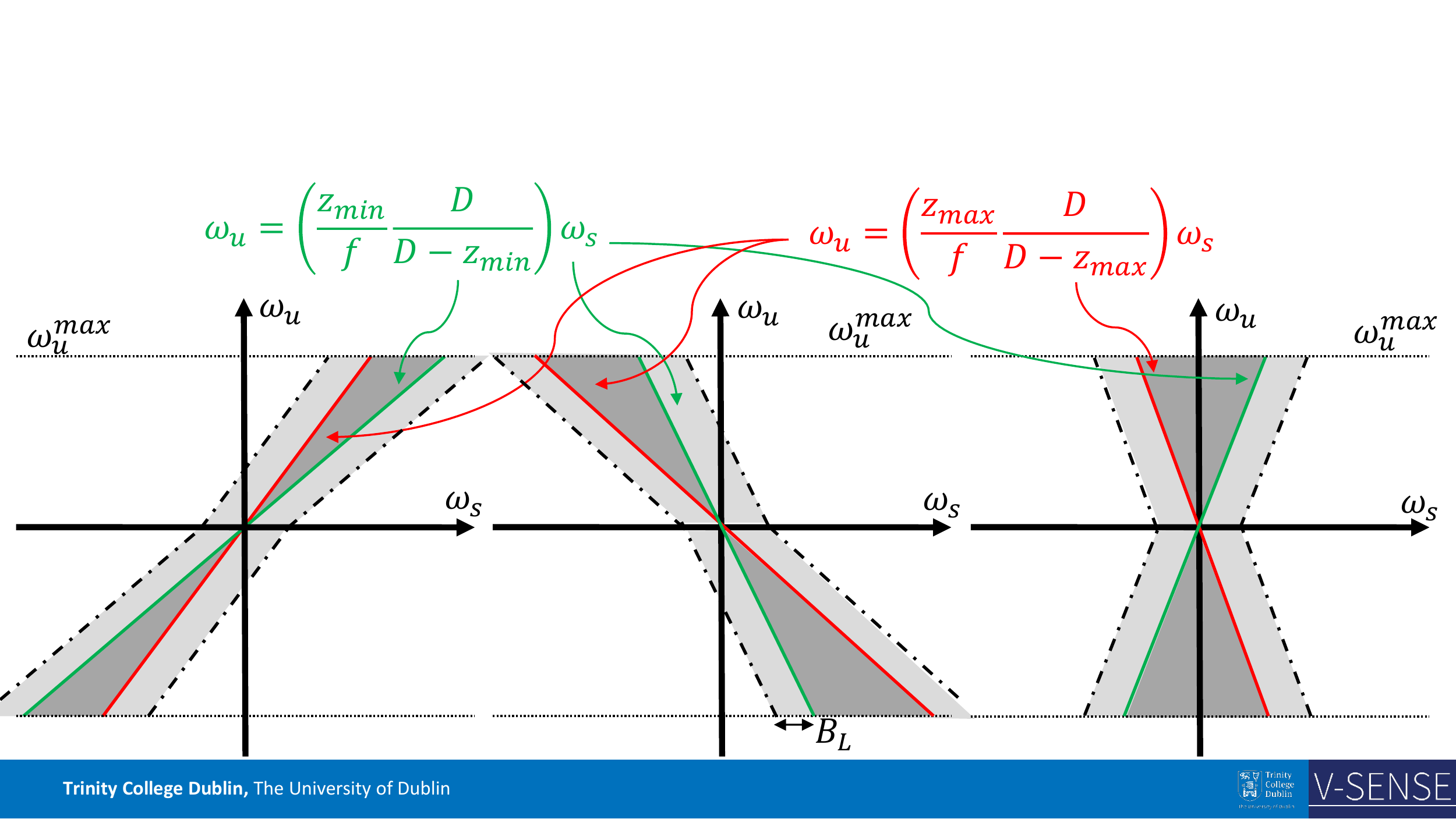}
	\centering
	\begin{tabular}{ccc}
	(a) $D > z_{max}$ & (b) $D < z_{min}$ & (c) $z_{min} < D < z_{max}$ \\
	\end{tabular}
	\caption{Bounds of the 2D light field spectrum in the Fourier domain depending on the parameterization of the image plane.}
	\label{fig:EPISpectrum}
\end{figure}

\begin{figure}
	\centering
	\begin{tabular}{ccc}
	Scene geometry & EPI & EPI Spectrum \\
	\includegraphics[width=0.25\linewidth,trim={0cm 0cm 0cm 0cm},clip]{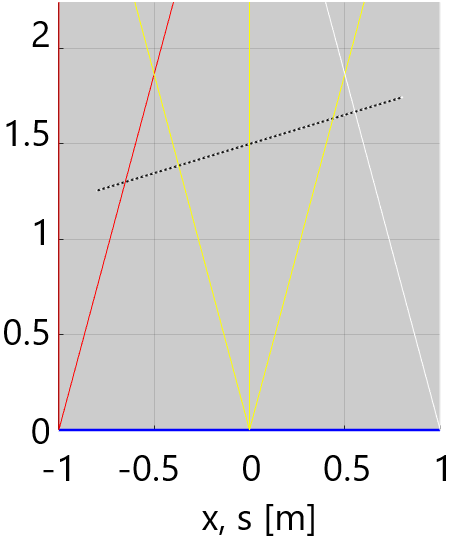} &
	\includegraphics[width=0.3\linewidth,trim={0cm 0cm 0cm 0cm},clip]{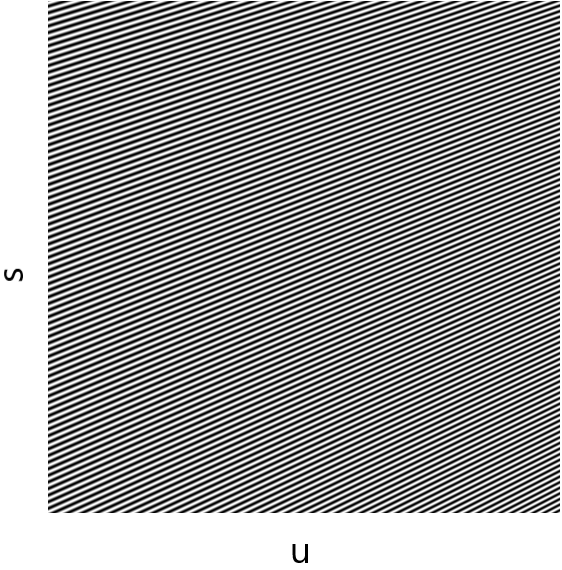} &
	\includegraphics[width=0.3\linewidth,trim={0cm 0cm 0cm 0cm},clip]{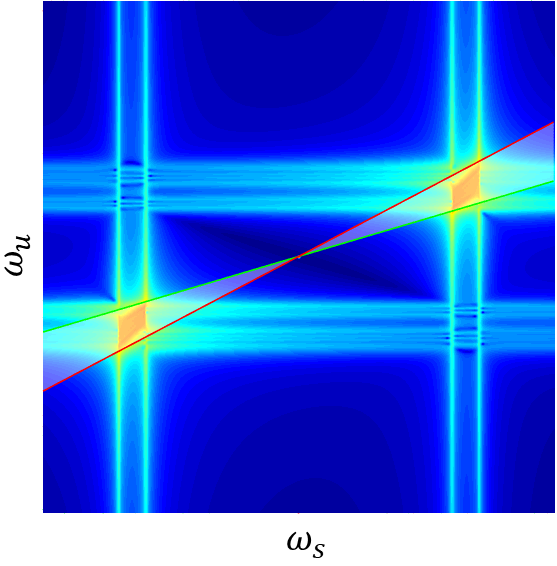} \\
	\multicolumn{3}{c}{(a) Image plane at infinity $D \to \infty$.} \\
	\includegraphics[width=0.25\linewidth,trim={0cm 0cm 0cm 0cm},clip]{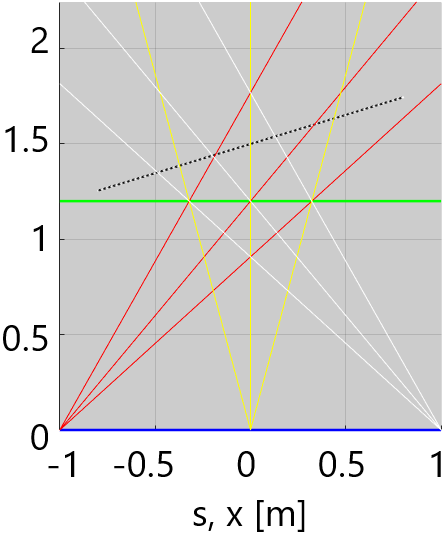} &
	\includegraphics[width=0.3\linewidth,trim={0cm 0cm 0cm 0cm},clip]{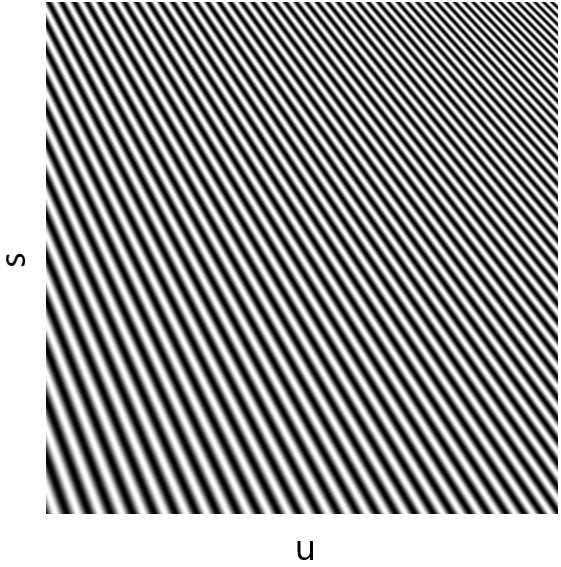} &
	\includegraphics[width=0.3\linewidth,trim={0cm 0cm 0cm 0cm},clip]{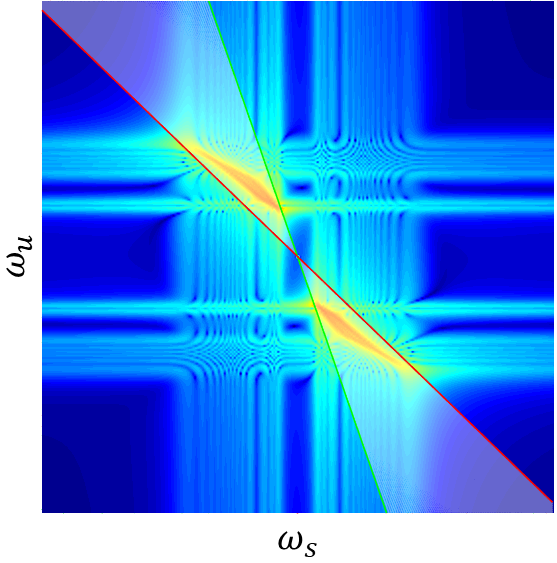} \\
	\multicolumn{3}{c}{(b) Image plane re-parameterized in front of the scene $D < z_{min}$.} \\
	\includegraphics[width=0.25\linewidth,trim={0cm 0cm 0cm 0cm},clip]{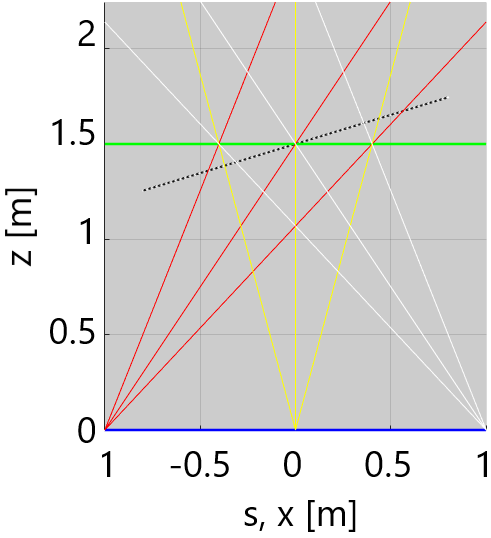} &
	\includegraphics[width=0.3\linewidth,trim={0cm 0cm 0cm 0cm},clip]{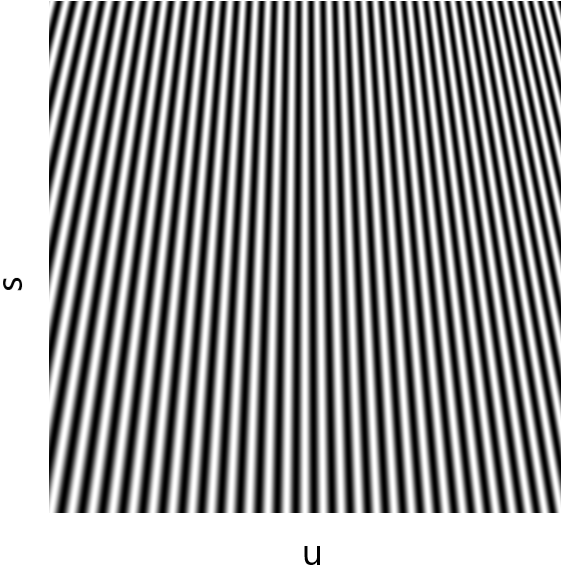} &
	\includegraphics[width=0.3\linewidth,trim={0cm 0cm 0cm 0cm},clip]{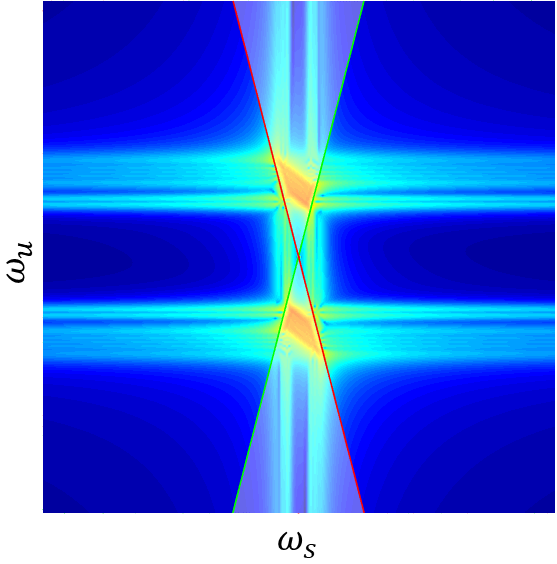} \\
	\multicolumn{3}{c}{(c) Image plane re-parameterized at the optimal depth $D = z_G$.} \\
	\end{tabular}
	\caption{Simulation results with two-parallel plane parameterization where the scene consists of a cosine texture pasted on a tilted plane. In the 2D scene representation, the camera plane is shown as a blue line, the image plane as a green line, camera frustums geometry are shown as red, yellow, and white lines, and the scene plane is shown as a tilted black and white line. The re-parameterization of the global image plane corresponds to a shearing of the EPI and its Fourier spectrum. Theoretical spectrum bounds (corresponding to Fig.\ref{fig:EPISpectrum}) are highlighted as green and red line on the spectrum images on the right. The EPI spectrum is most compacted along the spatial frequency axis $\omega_u$ when the image plane is parameterized at the optimal depth $z_G$ as defined in equation~\ref{eq:zG}.}
	\label{fig:tilted_plane_sin_texture_scene_TPP}

\end{figure}

\subsection{Angular sampling and reconstruction filters}

Assuming that the scene depth is bounded in the range $[z_{min},z_{max}]$, then the light field spectrum is bounded between two lines with slopes $\frac{z_{min}}{f}\frac{D}{D-z_{min}}$ and $\frac{z_{max}}{f}\frac{D}{D-z_{max}}$, giving it the well-known fan-shape shown in Fig.~\ref{fig:EPISpectrum}.
When the global image plane is at infinity $D \to \infty$, we obtain the exact same slopes as in previous papers, $\frac{z_{min}}{f}$ and $\frac{z_{max}}{f}$.
However, we have here a more generic expression which explicitly includes the global image plane parameter $D$.
Fig.~\ref{fig:EPISpectrum} shows different light field spectrum supports depending on the placement of the global image plane.
To obtain the optimal placement $D_{opt}$ corresponding to the maximum compaction of the spectrum along the spatial frequency axis $\omega_s$, we can see that we must solve the following equation:

\begin{equation*}
    \frac{z_{min}}{f}\frac{D_{opt}}{D_{opt}-z_{min}} = - \frac{z_{max}}{f}\frac{D_{opt}}{D_{opt}-z_{max}}
\end{equation*}

\noindent The solution to this equation gives:

\begin{equation}
\label{eq:Dopt}
    D_{opt} = \frac{2}{z_{min}^{-1} + z_{max}^{-1}}
\end{equation}

\noindent and the boundary line slopes are then equal to $\pm \frac{2z_{min}z_{max}}{f(z_{min} + z_{max})}$. This value is equal to $z_{opt}$ defined in equation~\ref{eq:zopt}, first derived in~\cite{chai2000plenoptic} as the optimal depth for a reconstruction filter with constant depth.
This means that we can replace this anti-aliasing non-separable reconstruction filter with constant depth by first re-parameterizing the global image plane at $D_{opt}$, and then applying a filter along the angular dimension only.
A similar observation was made in~\cite{do2011bandwidth} by looking at the shearing of the light field EPI.
The main difference in this paper is that we consider the EPI shearing as a consequence of the explicit (re-)parameterization of the global image plane, which gives us a new way of computing the optimal location $D_{opt}$ of the global image plane.

In Fig.~\ref{fig:tilted_plane_sin_texture_scene_TPP}, we further illustrate the effect of re-parameterizing the two-parallel plane light field, showing an example of results obtained with a scene object surface consisting of a tilted plane and a cosine texture (see section~\ref{sec:results} for a more detailed explanation of our experimental setups).
We can see that the results coincide with the expected theoretical fan-shape of the light field spectrum shown in Fig.~\ref{fig:EPISpectrum}.
Note that since we use a tilted plane as object surface, the optimal depth for the light field image plane is $z_G$, defined in equation~\ref{eq:zG}.

Considering that the highest spatial frequency $\omega_u^{max}$ is known, we can now also derive the maximum sampling interval $\Delta s_{max}$ along the angular dimensions based on the light field spectrum shape and the Nyquist sampling theorem.
From Fig.~\ref{fig:EPISpectrum}, we can see that the minimum interval to avoid any replica of the light field spectrum to overlap is:

\begin{multline}
    \label{eq:min_sampling_replica}
    \Delta\omega_s^{min} = \lvert \omega_s^{max} + B_L - (\omega_s^{min} - B_L) \rvert \\
    = \lvert \left( \frac{f}{z_{min}}\frac{D-z_{min}}{D} \right) \omega_u^{max}  - \left( \frac{f}{z_{max}}\frac{D-z_{max}}{D} \right)  \omega_u^{max} + 2B_L\rvert \\
    = \lvert \left(\frac{D z_{max} - z_{min}z_{max} - (D z_{min} - z_{max}z_{min}) }{D z_{max} z_{min}}\right) f \omega_u^{max} + 2B_L \rvert\\
    = \lvert \frac{D (z_{max} - z_{min}) }{D z_{max} z_{min}} f \omega_u^{max}+ 2B_L \rvert\\
    = \lvert (z_{min}^{-1} - z_{max}^{-1}) f \omega_u^{max} + 2B_L \rvert \\
\end{multline}

\noindent and thus, we have:

\begin{equation}
    \label{eq:smax_bl}
    \Delta s_{max} = \frac{1}{\Delta\omega_s^{min}} 
    =  \lvert f(\frac{1}{z_{min}} - \frac{1}{z_{max}}) \omega_u^{max} + 2B_L \rvert^{-1}
\end{equation}

\noindent which is similar to the result of equation~\ref{eq:smax} but takes into account the light field spectrum expansion due to non-Lambertian properties of the surface light field.

\section{Spectral analysis of the tilted image plane light field parameterization}
\label{sec:TIP} 

\begin{figure}[t]
	\centering
	\includegraphics[width=\linewidth,trim={0cm 1.1cm 0cm 0cm},clip]{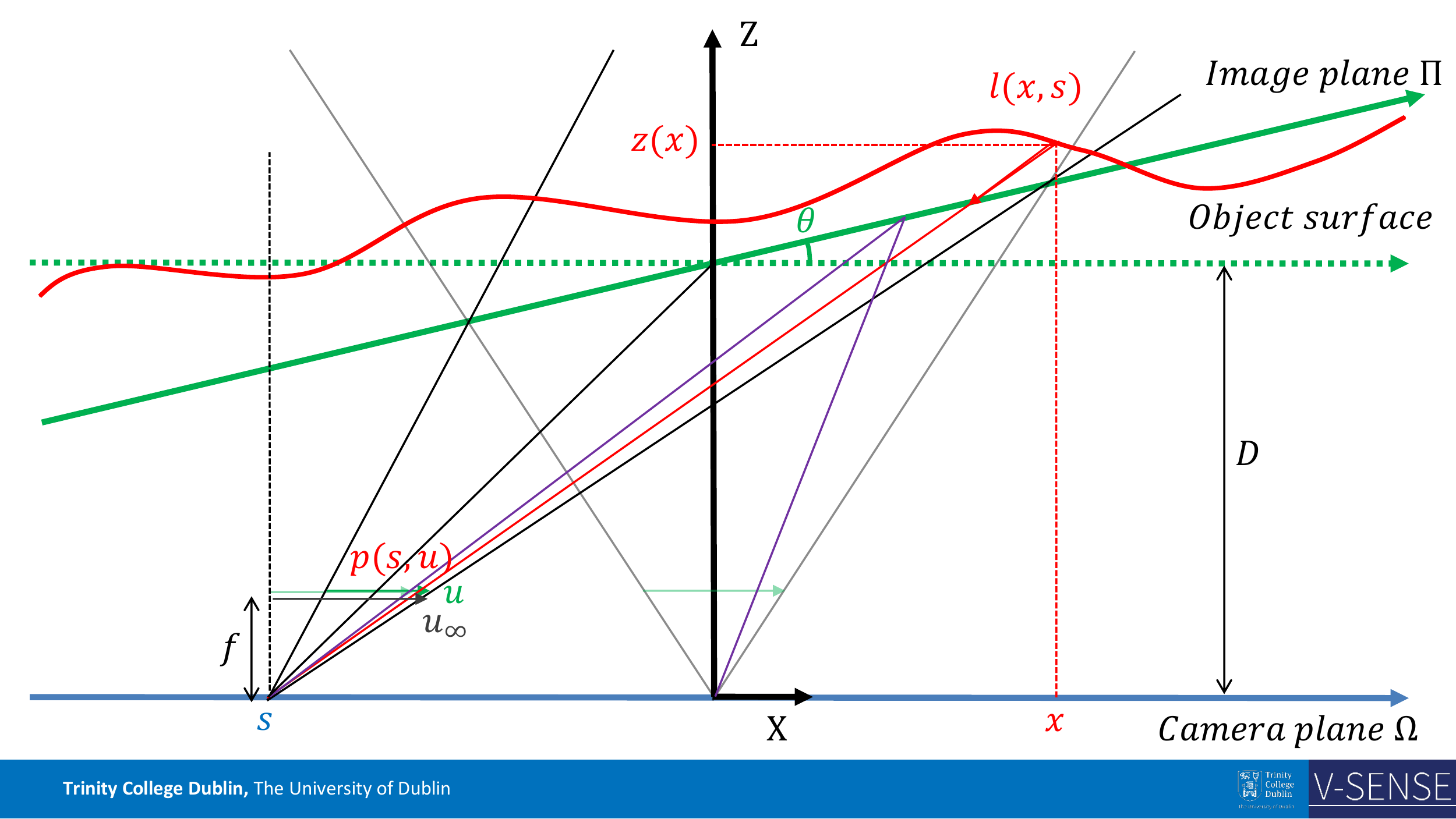}
	\caption{Tilted Image Plane parameterization for a 2D slice  of the light field.}
	\label{fig:TIP}
\end{figure}

In this section, we further extend the re-parameterization capabilities of the two plane light field by allowing to tilt the image plane.

\subsection{Geometric mapping}

We use here the same 2D scene model as in previous section on the two-parallel plane light field, but consider that the image plane, in addition to being positioned at depth $D$, is tilted with an angle $\theta$ (see Fig.~\ref{fig:TIP}).
The first consequence of tilting the image plane is that the light field camera position is bounded by the intersection of the image plane with the camera plane, which position we denote as:

\begin{equation}
    \label{eq:max_s_tilt}
    s_{\theta}=-\frac{D}{\tan(\theta)}
\end{equation}

The expression linking the local image plane coordinate $u$ to $u_\infty$, corresponding to the previous equation~\ref{eq:u_inf}, is then:

\begin{equation}
    \label{eq:u_inf_tilt}
    u = \left(1 + \frac{s\tan(\theta)}{D} \right)\left(u_{\infty} + \frac{sf}{D}\right)
\end{equation}

In addition, the geometry of the light rays expressed in equation~\ref{eq:lf_geo_u_inf} is still valid, and by combining it with equation~\ref{eq:u_inf_tilt} (see appendix~\ref{sec:TIP_app} for more details), we obtain the generic geometric mapping linking a light ray $(s,u)$ to its point of origin on the object surface at position $(x, z(x))$:

\begin{equation}
    \label{eq:lf_geo_tilt}
    u = \left(1 + \frac{s\tan(\theta)}{D}\right)\left(\frac{xf}{z(x)} + s f \left(\frac{1}{D} - \frac{1}{z(x)}\right)\right)
\end{equation}

Note that in practice, when generalizing to a full 4D light field, the re-parameterization of a parallel image plane through a tilted image plane can be interpreted as the projection of the image coordinate from the centre camera (i.e. camera at position $s=0,t=0$) to a camera at position $s,t$ through the tilted plane, and can be implemented using homographies.

\subsection{Spectral analysis}

Using the new geometric mapping of equation~\ref{eq:lf_geo_tilt}, we can carry an analysis of the light field spectrum $P$ by following a similar reasoning as in the previous section (see appendix~\ref{sec:TIP_app} for the full derivation), which leads to the following result:

\begin{equation}
\label{eq:P_fourier_tr_tilt}    
    P(\omega_s, \omega_u) = \int_{-\infty}^{\infty}  \frac{f}{z(x)^2} e^{-j\left(\omega_u \frac{xf}{z(x)}\right)} \left(H_\theta(x, \omega_s) * C(x, \omega_s)\right) dx
\end{equation}

\noindent where we define:

\begin{equation*}
    h_\theta(x,s) \triangleq (1 + \frac{s\tan(\theta)}{D})(z(x) + z'(x) (s-x)) l(x,s)    
\end{equation*}
\begin{equation*}
    c(x,s) \triangleq e^{j\left((\omega_u f (\frac{1}{z(x)} - \frac{1}{D} - \frac{\tan(\theta)x}{D z(x)})) s + \omega_u f \frac{\tan(\theta)}{D} (\frac{1}{z(x)} - \frac{1}{D}) s^2 \right)}
\end{equation*}

\noindent with $H_\theta(x, \omega_s)$ and $C(x, \omega_s)$ the Fourier transform of $h_\theta$ and $c$ over $s$ respectively, and $*$ refers to the convolution over $\omega_s$.

The function $c$ corresponds to a \textit{chirp} function, which has found applications in frequency modulation for telecommunications and radar systems~\cite{klauder1960theory}, where the frequency of the signal increases or decreases with time.
Here, $c$ is a linear chirp function of the form  $e^{j\left(\omega_0 s + \lambda_c s^2\right)}$ where the linear frequency sweep is centred on $\omega_0$ and increases or decreases with slope 
$\lambda_c$, such that:

\begin{equation*}
    \omega_0 = \frac{\omega_u f}{Dz(x)} (D-z(x)-\tan(\theta)x)
\end{equation*}

\begin{equation*}
    \lambda_c = \frac{\omega_u f}{Dz(x)} \frac{\tan(\theta)(D-z(x))}{D}
\end{equation*}

By denoting $\phi_c(s) = \omega_0 s + \lambda_c s^2$ the phase of the linear chirp, we can express the ``instantaneous'' frequency of the chirp as:

\begin{equation*}
    \omega_c(s) = \frac{d \phi_c}{ds} =  \omega_0 + 2\lambda_c s
\end{equation*}

By recalling that $s$ is bounded by $s_\theta$ as defined in equation~\ref{eq:max_s_tilt}, we can then derive an upper bound of the chirp bandwidth as:

\begin{multline*}
    B_C = \omega_c(s_\theta) 
    = \frac{\omega_uf}{Dz(x)} (D - z(x) - \tan(\theta)x + 2 \frac{\tan(\theta)(D-z(x))}{D} \frac{-D}{\tan(\theta)}) \\
    = \frac{\omega_uf}{Dz(x)} \left(D - z(x) - \tan(\theta)x - 2(D-z(x) \right) \\
    = \frac{\omega_uf}{Dz(x)} \left(z(x) - D - \tan(\theta)x \right) \\
\end{multline*}

In addition, using the Fourier transform properties, we have:

\begin{multline}
\label{eq:H_fourier_tr_tilt}
    H_\theta(x,\omega_s) = 
    \left(1 + \frac{s\tan(\theta)}{D}  \right) (z(x) - x z'(x)) L(x, \omega_s) + \\ j (z'(x) + \frac{\tan(\theta)}{D}(z(x)-x z'(x)) ) \frac{\partial L(x, \omega_s)}{\partial\omega_s} - \frac{\tan(\theta)}{D} z'(x) \frac{\partial^2 L(x, \omega_s)}{\partial^2\omega_s} \\
\end{multline}

We then follow a similar reasoning as in section~\ref{sec:TPP}, first recalling that the surface light field $l$ is band-limited over $s$ with bandwidth $B_L$, formally expressed in equation~\ref{eq:L_band_limit}.
The no self-occlusion condition of equation~\ref{eq:no_self_occ} ensures that we have $\left(1 + \frac{s\tan(\theta)}{D}  \right)(z(x) - x z'(x)) > 0$, which means that $B_L$ is an upper bound of the bandwidth of $H_\theta(x,\omega_s)$.
We can then consider that the bandwidth of $\left(H_\theta(x, \omega_s) * C(x, \omega_s)\right)$ is the sum of the bandwidths of $H_\theta$ and $C$, $B_L$ and $B_C$ respectively.
The essential bandwidth of the light field spectrum $P$ for the tilted image plane parameterization can thus be expressed as:

\begin{equation}
\label{eq:P_band_limit_tilt}
        P(\omega_s, \omega_u) = 0, \text{ if } \lvert \omega_s + \omega_u f \left(\frac{z(x) - D - \tan(\theta)x}{D z(x)}\right) \rvert > B_L
\end{equation}

\subsection{Angular sampling}

We now assume, without loss of generality, that the scene surface is parameterized as follows:

\begin{equation}
    z(x) = z_O + \tan(\theta_O)x + r(x)
\end{equation}

\noindent such that $r_{min} \leq r(x) \leq r_{max}$, i.e. the scene object surface is centred around a tilted plane within a bounded range of $[r_{min}, r_{max}]$.
If the surface is known, the parameters $z_O $ and $\theta_O$ can be estimated in practice using line fitting (or plane fitting when generalizing to a 3D scene).
Then, by parameterizing the light field image plane with $D=z_O$ and $\theta =\theta_O$, we can see that the light field spectrum is bounded by two lines with slopes $\frac{z_{min}}{f}\frac{D}{r_{min}}$ and $\frac{z_{max}}{f}\frac{D}{r_{max}}$.
Following a similar reasoning as used to derive the previous sampling guideline of equation~\ref{eq:smax_bl}, we obtain a new guideline for the camera spacing:

\begin{equation}
    \label{eq:smax_bl_tilt}
    \Delta s_{max}^\theta  = \lvert \frac{f}{z_O}\lvert\frac{r_{min}}{z_{min}} - \frac{r_{max}}{z_{max}}\rvert \omega_u^{max} + 2B_L \rvert^{-1}
\end{equation}

\begin{figure}
	\centering
	\small
	\begin{tabular}{ccc}
	Scene geometry & EPI & \multicolumn{1}{c}{EPI Spectrum }  \\ 
	\includegraphics[width=0.3\linewidth,trim={0cm 0cm 0cm 0cm},clip]{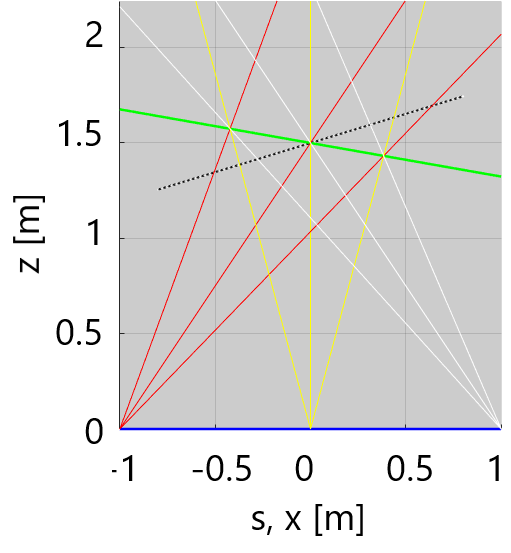} &
	\includegraphics[width=0.3\linewidth,trim={0cm 0cm 0cm 0cm},clip]{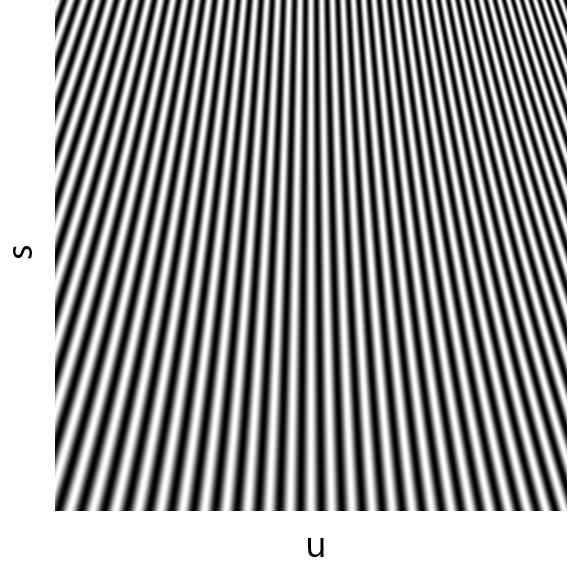} &
	\includegraphics[width=0.3\linewidth,trim={0cm 0cm 0cm 0cm},clip]{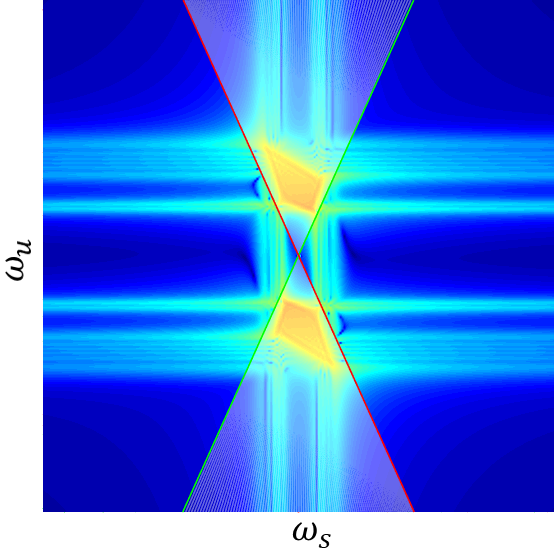} \\
	\multicolumn{3}{c}{(a) Image plane re-parameterized at an angle smaller than the scene plane angle $\theta < \theta_{O}$.} \\ 
	\includegraphics[width=0.3\linewidth,trim={0cm 0cm 0cm 0cm},clip]{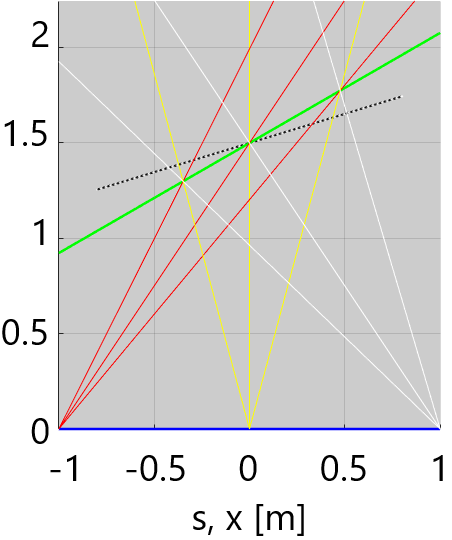} &
	\includegraphics[width=0.3\linewidth,trim={0cm 0cm 0cm 0cm},clip]{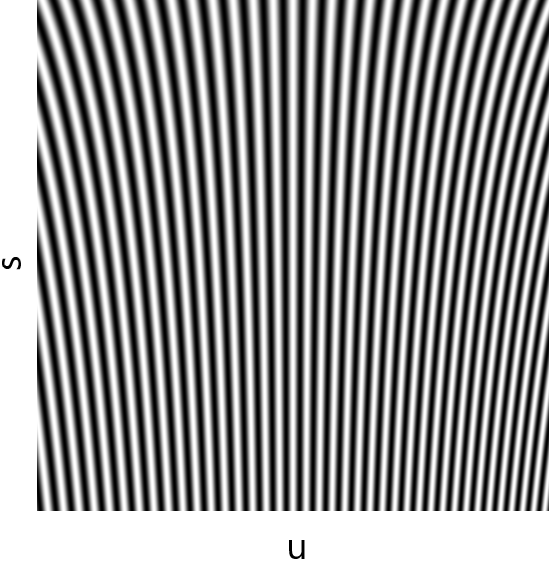} &
	\includegraphics[width=0.3\linewidth,trim={0cm 0cm 0cm 0cm},clip]{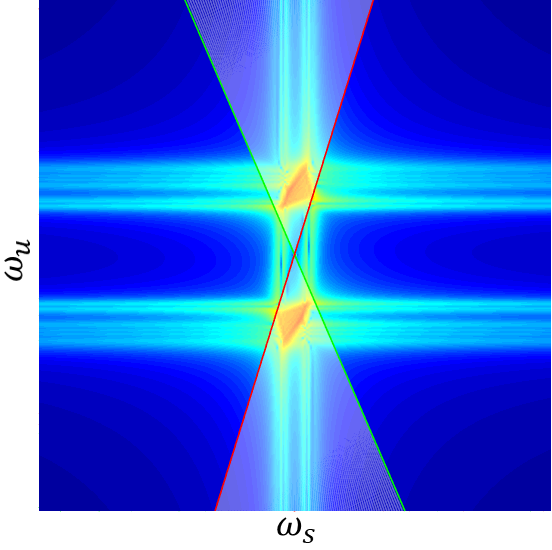} \\
	\multicolumn{3}{c}{(b) Image plane re-parameterized at an angle greater than the scene plane angle $\theta > \theta_{O}$.} \\ 
	\includegraphics[width=0.3\linewidth,trim={0cm 0cm 0cm 0cm},clip]{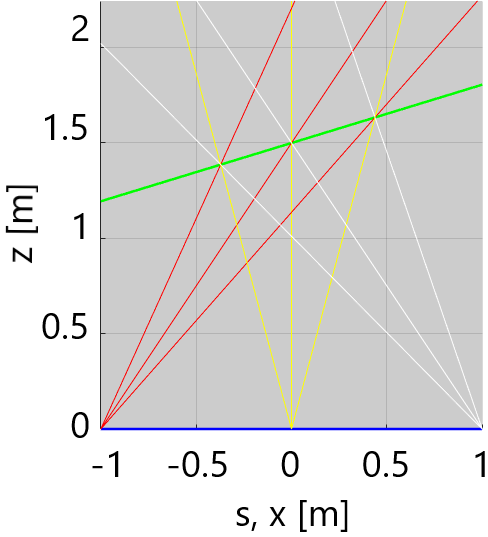} &
	\includegraphics[width=0.3\linewidth,trim={0cm 0cm 0cm 0cm},clip]{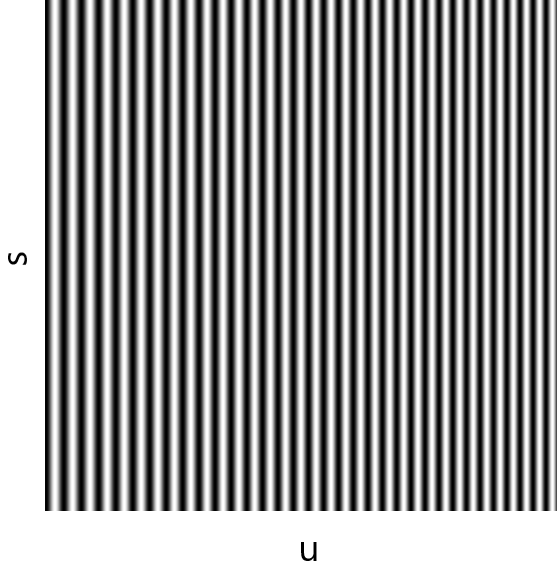} &
	\includegraphics[width=0.3\linewidth,trim={0cm 0cm 0cm 0cm},clip]{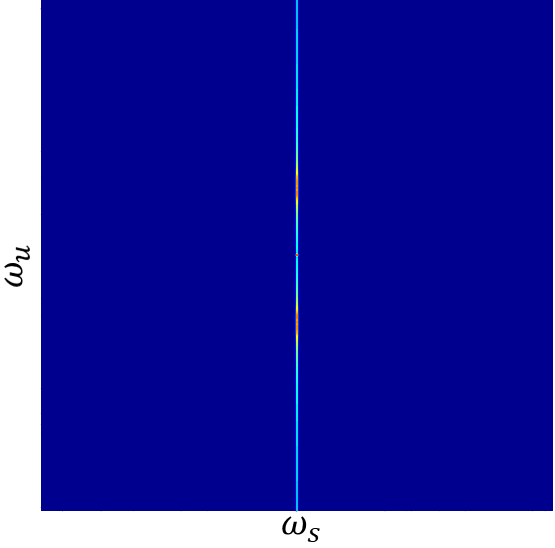} \\
	\multicolumn{3}{c}{(c) Image plane  re-parameterized at an angle equal to the scene plane angle $\theta = \theta_{O}$.} \\
	\end{tabular}
	\caption{Simulation results with tilted image plane parameterization where the scene consists of a cosine texture pasted on a tilted plane. The EPI spectrum is most compacted along the spatial frequency axis $\omega_u$ when the image plane coincides with the object surface. This additional degree of freedom in the image plane parameterization allows to obtain the most compact spectrum support compared to the two-parallel plane parameterization shown in Fig.~\ref{fig:tilted_plane_sin_texture_scene_TPP}.}

	\label{fig:tilted_plane_sin_texture_scene_TIP}
\end{figure}

A first observation is that if the scene object surface is exactly a tilted plane, i.e. $\forall x,\,r(x)=0$, then in particular $r_{min}=r_{max}=0$ and the slope of the lines bounding the light field spectrum tend to infinity, which means we achieve the maximum compaction of the spectrum along the spatial frequency axis $\omega_s$, and the largest camera spacing possible $\Delta s_{max}^\theta  = \lvert2B_L \rvert^{-1}$.
This is illustrated in Fig.~\ref{fig:tilted_plane_sin_texture_scene_TIP}, which shows an example of results obtained by re-parameterizing the angle $\theta$ of the light field image plane, where the scene object surface consists of a tilted plane and a cosine texture (as in Fig.~\ref{fig:tilted_plane_sin_texture_scene_TPP}).
However, when the image plane orientation is further away from the scene object orientation, we can see that the spectrum support actually expands.

More generally, we can see that if the range $[r_{min}, r_{max}]$ is such that  $\frac{1}{z_O}(\frac{r_{min}}{z_{min}} - \frac{r_{max}}{z_{max}}) \leq \frac{1}{z_{min}} - \frac{1}{z_{max}}$, then the new sampling guideline of equation~\ref{eq:smax_bl_tilt} is more advantageous than the previous guideline given in equation~\ref{eq:smax_bl} for the two-parallel plane light field.
\section{Simulation results}
\label{sec:results}

\subsection{Experimental setup}

We use in our experimentations synthetic 2D scenes and light fields rendered in the $(x,z)$ plane.
Since our goal is to study the spectral properties of the light field and its sampling, we need access to reference dense light fields with ground truth texture and geometry. Existing dense light field datasets of natural scenes such as~\cite{CIVITdataset} do not provide the ground truth geometry, while existing synthetic datasets providing ground truth depth maps such as~\cite{honauer2016dataset} are not dense enough, i.e. do not provide the ground truth texture as it is already sub-sampled. Thus, following exiting work studying light field sampling~\cite{zhang2003spectral,do2011bandwidth,gilliam2013spectrum}, we model both the scene surface and texture using an explicit analytical expression, and choose the scene and texture parameters such that the rendered light fields do not exhibit any aliasing. 

Each scene contains a single object consisting of a tilted plane (line in 2D) or quadratic surface, such that the surface of the object can be expressed using the following equation:
\begin{equation}
    \label{eq:test_geo}
    z(x)=z_O + \tan(\theta_O)x + qx^2
\end{equation}
The object Lambertian texture $\Lambda(x)$ consists of a combination of cosine waves with different frequencies, expressed as:
\begin{equation}
    \label{eq:test_tex}
    \Lambda(x)=\frac{1}{2K}\sum_{k=1}^K(\cos(\omega_k x)+1)
\end{equation}

\begin{figure}
	\centering

	\begin{tabular}{cc}
	\includegraphics[width=0.4\linewidth,trim={0cm 0cm 0cm 0cm},clip]{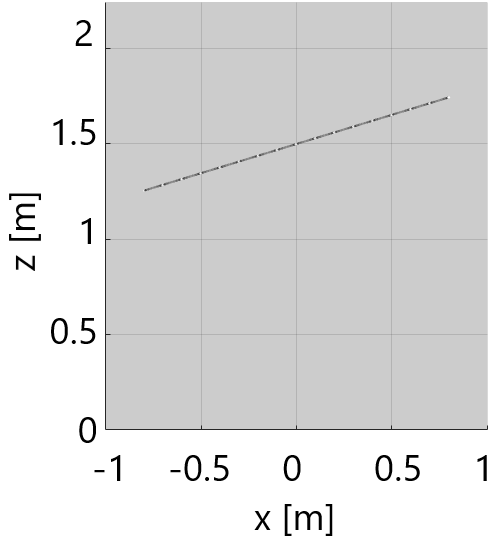} &
	\includegraphics[width=0.4\linewidth,trim={0cm 0cm 0cm 0cm},clip]{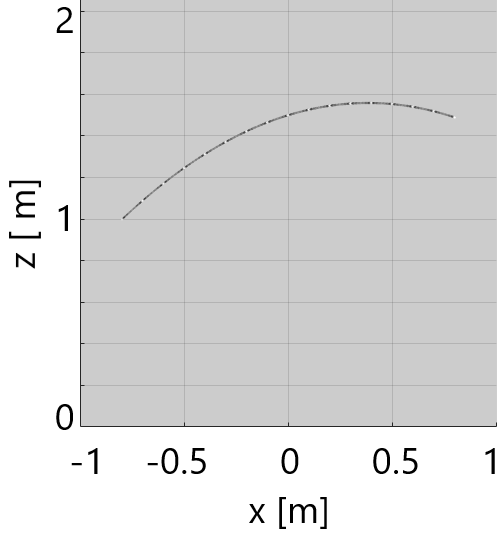} \\
	Scene A & Scene B \\
	\includegraphics[width=0.4\linewidth,trim={0cm 0cm 0cm 0cm},clip]{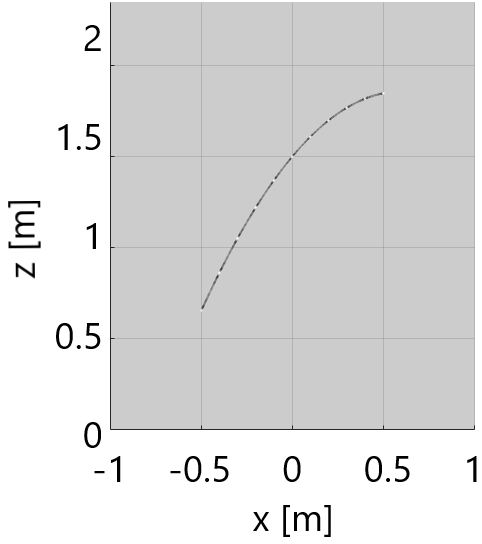} &
	\includegraphics[width=0.5\linewidth,trim={0cm 0cm 0cm 0cm},clip]{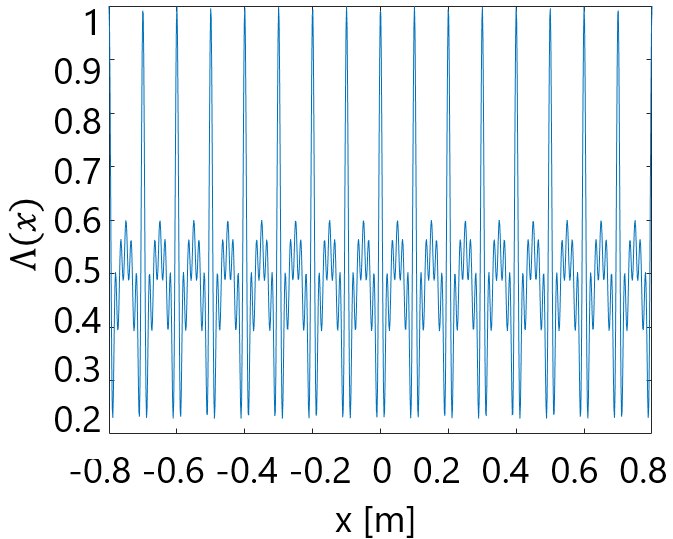} \\
	Scene C & Lambertian texture $\Lambda$ \\
	\end{tabular}
	\caption{Geometry and texture of the 2D synthetic scenes used in our experiments.}

	\label{fig:scenes}

\end{figure}

Note that the texture is monochrome and normalized to the range $[0, 1]$.
Test light fields are then rendered by projecting rays from each camera position $s$ (which corresponds to a line in the EPI) and determine their intersections with the scene surface using the analytical expression of equation~\ref{eq:test_geo}. The $x$ coordinate of the intesection points are used to query the texture values using equation~\ref{eq:test_tex}.
We create 3 different scenes, denoted A, B and C.
The geometry of the object in scene A is a tilted plane, while for scenes B and C it is a quadratic curve.
The texture of all scenes is composed of $K=5$ different frequencies.
This is illustrated in Fig.~\ref{fig:scenes}.
The exact scene parameters used in our experiments are summarized in Table~\ref{tab:planar_scene_param}.
Note that scene C has been designed specifically for the last experiment in which we divide the scene geometry in multiple non-overlapping depth layers, and thus requires a larger depth range. For all other experiments, we use scenes A and B.
The results shown in Figs.~\ref{fig:tilted_plane_sin_texture_scene_TPP} and~\ref{fig:tilted_plane_sin_texture_scene_TIP} are obtained using a variant of scene A with a single cosine texture where $\omega_1 = 40.$

For simplicity, the focal length of the light field camera is set to $f=1$.
The camera plane boundary is set to $s_{max} = 1$.
The local image plane boundary is set to $u_{max} = 0.2679$, which corresponds to a field of view of $30^\circ$.
The light field EPIs are rendered with a resolution of $512\times512$ pixels unless specified otherwise. 
\begin{table}[]
    \centering
    \begin{tabular}{|c|c|c|c|} 
    \hline
    Scene parameters & A & B & C \\ 
    \hline
        $[x_{min}, x_{max}]$ [m] & \multicolumn{2}{c|}{[-0.8, 0.8]} & [-0.5, 0.5]       \\
                $z_O$ [m]              & \multicolumn{3}{c|}{1.5}       \\
        $\theta_O$ [$^\circ$]  & \multicolumn{2}{c|}{17} & 50       \\
        $q$ [m$^{-1}$]         & 0         & -0.4   & -1\\
        $z_{min}$ [m]          & 1.2554    & 0.9994 & 0.6541\\
        $z_{max}$ [m]          & 1.7446    & 1.5584 & 1.8459\\
        $\omega_k$ [rad/m] (K=5)     & \multicolumn{3}{c|}{20, 30, 40, 50, 60}  \\ 
    \hline
    \end{tabular}
    \caption{Parameters used to generate the 2D synthetic scenes used in our simulations.}
    \label{tab:planar_scene_param}
\end{table}

\begin{figure}[t!]
	\centering
	\begin{tabular}{cc}
	\multicolumn{2}{c}{Scene A} \\
	\includegraphics[width=0.5\linewidth,trim={0cm 0cm 0cm 0cm},clip]{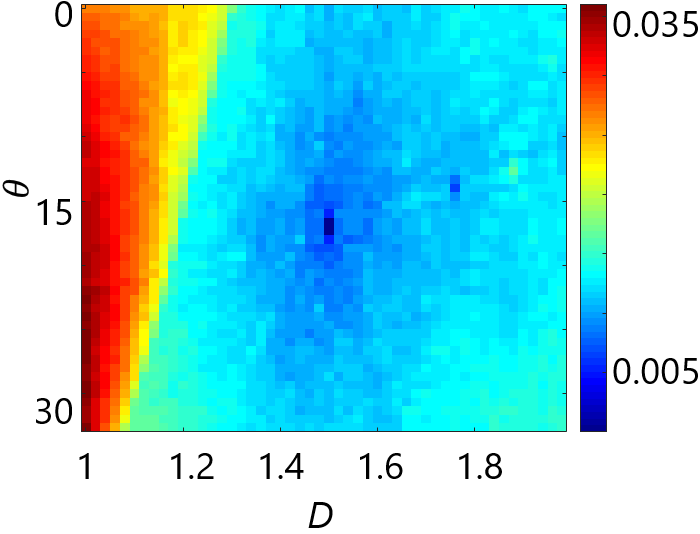} &
	\includegraphics[width=0.5\linewidth,trim={0cm 0cm 0cm 0cm},clip]{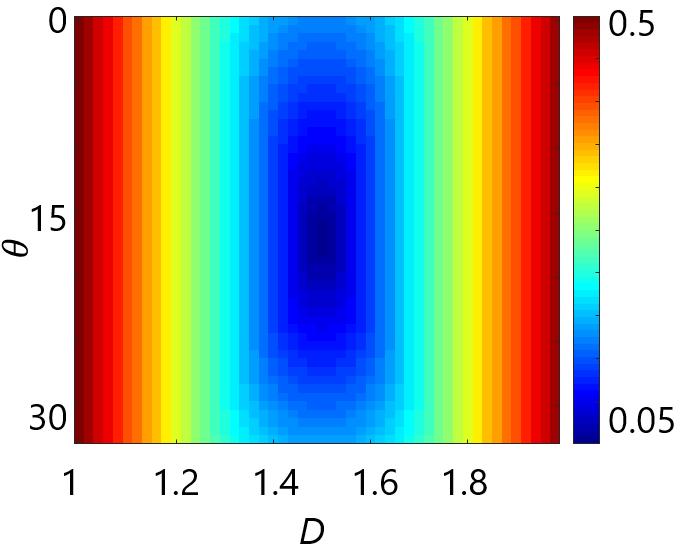} \\
	\multicolumn{2}{c}{Scene B} \\
	\includegraphics[width=0.5\linewidth,trim={0cm 0cm 0cm 0cm},clip]{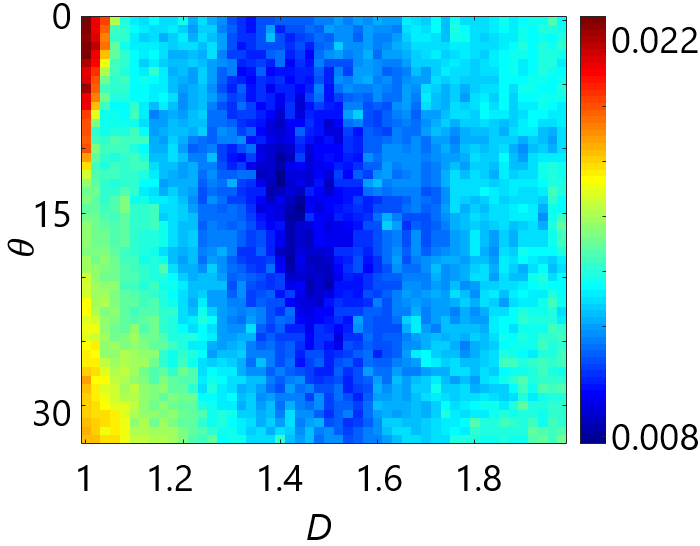} &
	\includegraphics[width=0.5\linewidth,trim={0cm 0cm 0cm 0cm},clip]{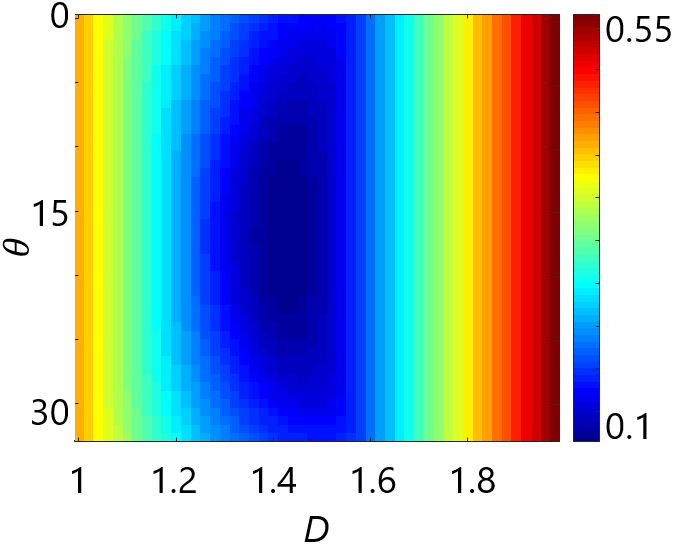} \\
	\end{tabular}
	\caption{Measure of the sparsity of the spectrum (RMSE of decimated spectrum, lower is better) depending on the image plane depth $D$ and orientation $\theta$ shown on the left for scene A (top) and B (bottom). The corresponding distance between the image plane and scene geometry is shown on the right. For both scenes, a clear minimum (maximum sparsity) is reached when the image plane best fit the scene geometry. (Note that the colour range is different for each subfigure in order to clearly show the minimum, please refer to the colorbar to see the error range.)}
	\label{fig:spectrum_sparse_opt_vs_geo}
\end{figure}

\subsection{Evaluation of the re-parameterized light field spectrum sparsity}

As observed in Fig.~\ref{fig:tilted_plane_sin_texture_scene_TIP}, the light field image plane parameterization will impact the compactness of the light field spectrum support, with more compact spectrum allowing to relax the sampling guidelines.
In our first experiment, we propose to further study the compactness of the light field spectrum support by evaluating the light field spectrum sparsity for a range of image plane position $D$ and orientation $\theta$.
We use 50 different values for $D$, equally spaced between 1 and 2 meters, and 50 different values for $\theta$, equally space between 0 and $2\times\theta_O = 34^\circ$.

To measure the sparsity, we compute the Root Mean Square Error (RMSE) between the light field spectrum and a decimated version of the spectrum containing only the 1\% highest coefficients of the original spectrum, such that a lower RMSE indicates a sparser spectrum.
We use for this experiment scenes A and B, and the results are shown in Fig.~\ref{fig:spectrum_sparse_opt_vs_geo}.
We also show the mean absolute error (MAE) between the light field image plane and the scene object surface as a measure of the image plane fit with the scene geometry.
For both scenes, the spectrum is the most sparse when the image plane best fits the scene geometry.

\begin{figure}[t!]
	\centering
	\begin{tabular}{ccc}
	\multicolumn{3}{c}{Scene A} \\
	\includegraphics[width=0.3\linewidth,trim={0cm 0cm 0cm 0cm},clip]{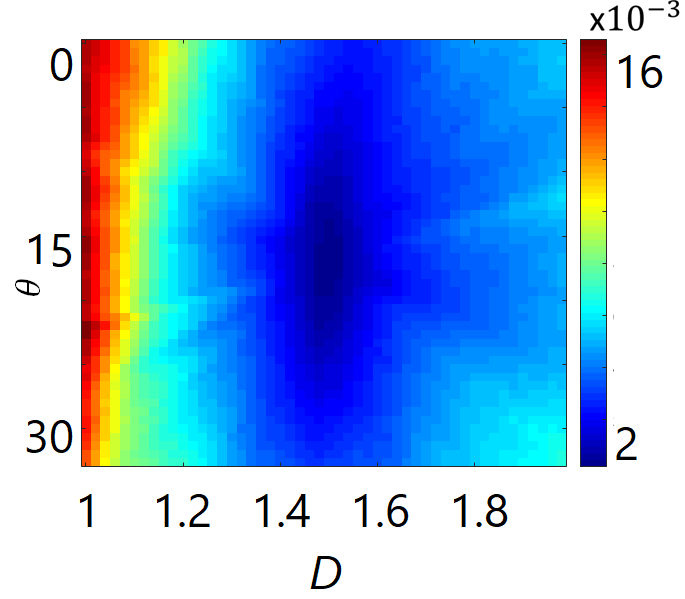} &
	\includegraphics[width=0.3\linewidth,trim={0cm 0cm 0cm 0cm},clip]{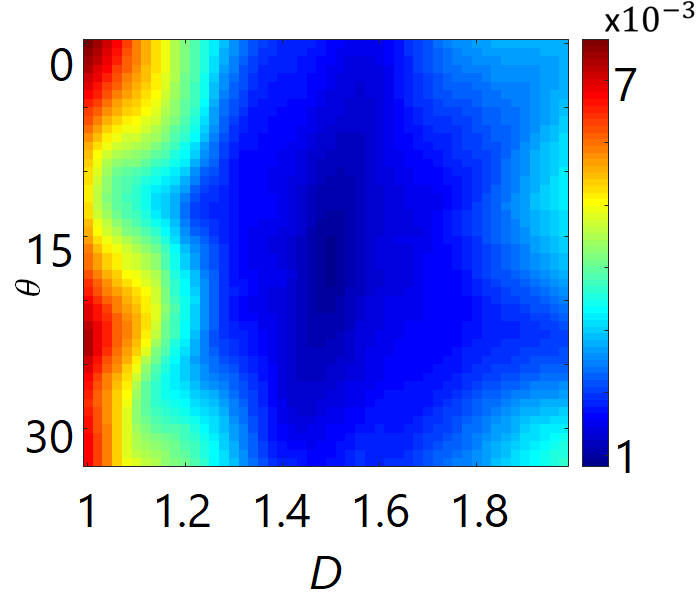} &
	\includegraphics[width=0.3\linewidth,trim={0cm 0cm 0cm 0cm},clip]{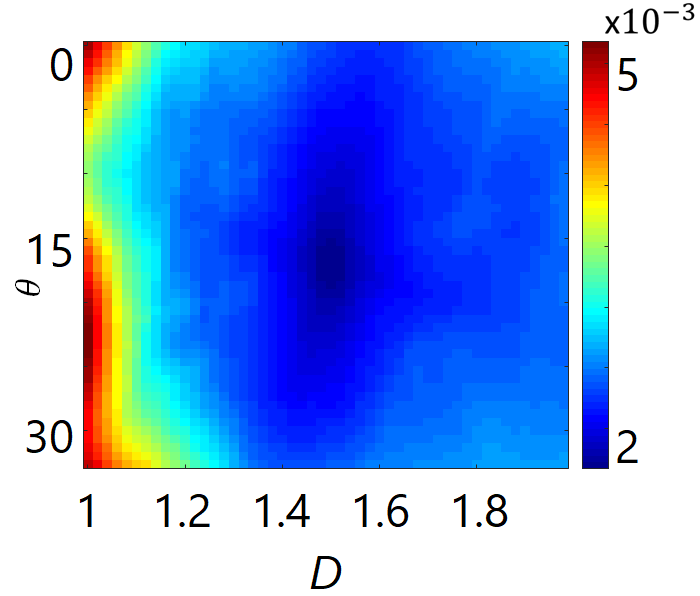} \\

    \multicolumn{3}{c}{Scene B} \\
	\includegraphics[width=0.3\linewidth,trim={0cm 0cm 0cm 0cm},clip]{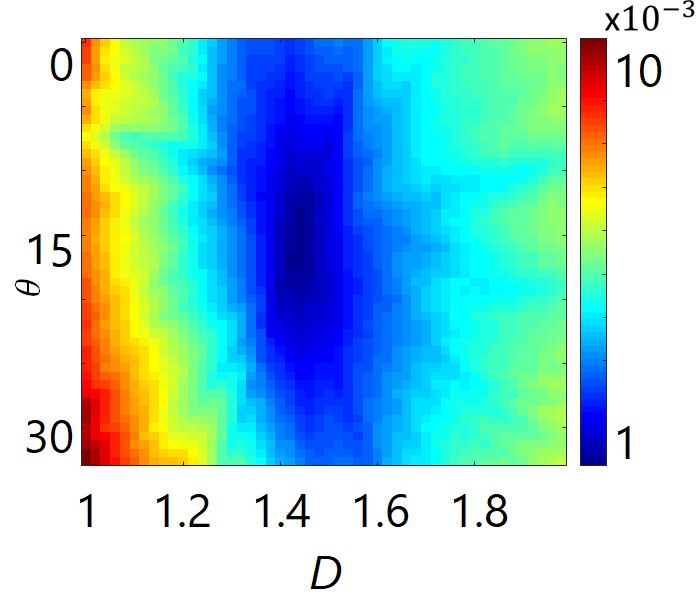} &
	\includegraphics[width=0.3\linewidth,trim={0cm 0cm 0cm 0cm},clip]{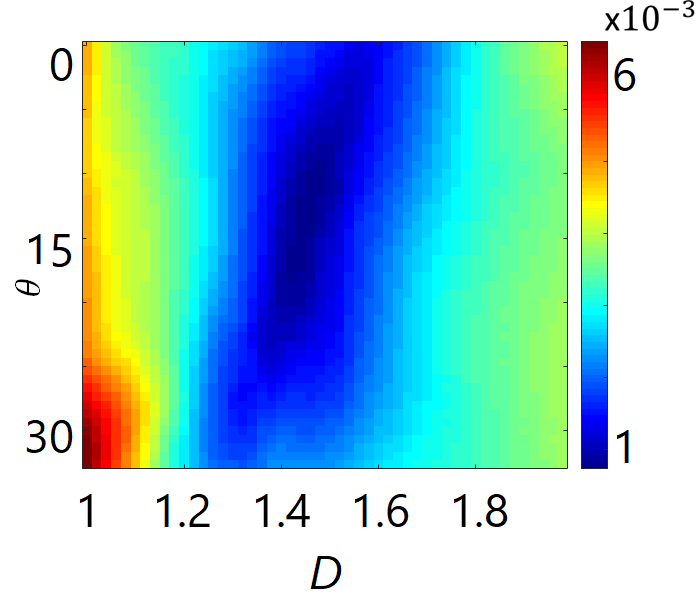} &
	\includegraphics[width=0.3\linewidth,trim={0cm 0cm 0cm 0cm},clip]{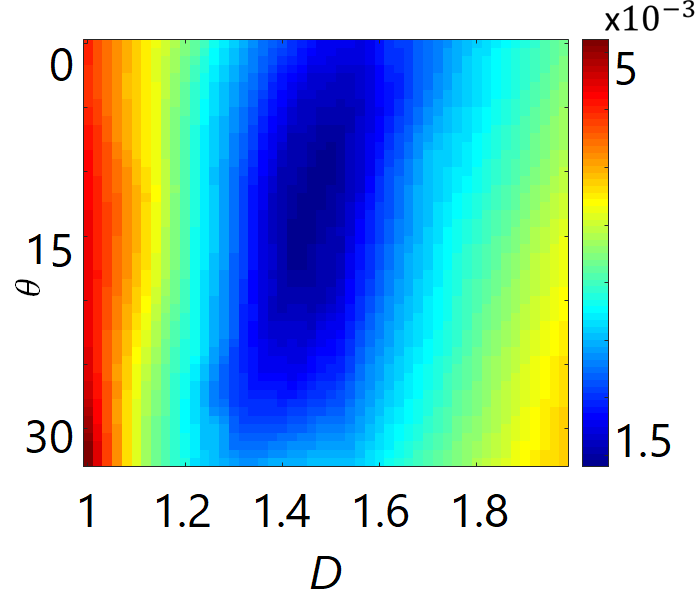} \\
	$B_L=1$ & $B_L=5$ & $B_L=10$ \\
	\end{tabular}
	\caption{Measure of the sparsity of the spectrum (RMSE of decimated spectrum, lower is better) depending on the image plane depth $D$ and orientation $\theta$ for scene A (top) and B (bottom), using from left to right a non-Lambertian texture with $B_L = 1, 5, 10$. The minima observed in our previous experiment shown in Fig.~\ref{fig:spectrum_sparse_opt_vs_geo} are still clearly visible.}
	\label{fig:sparse_opt_vs_Bl}
\end{figure}

To further test the robustness of this observation, we propose in our second experiment to conduct a similar study, but using a non-Lambertian texture $\Gamma(x,s)$, which we define as:

\begin{equation}
    \Gamma(x,s)=\Lambda(x)\sinc(B_L s)
\end{equation}

\noindent where the sinc function $\sinc(B_L s)$ is used to simulate a bandlimited non-lambertian texture with bandwidth $B_L$.
We can expect the previous observation to be confirmed, as the bandwidth of the non-Lambertian texture should expand the light field spectrum support similarly for all light field parameterizations.
The results for both scenes A and B are shown in Figs.~\ref{fig:sparse_opt_vs_Bl}, and we can see that 
the minima observed in the previous experiment are still clearly visible. 

\begin{figure}
	\centering
	\begin{tabular}{ccc}
	\multicolumn{3}{c}{Scene A} \\
	\includegraphics[width=0.3\linewidth,trim={0cm 0cm 0cm 0cm},clip]{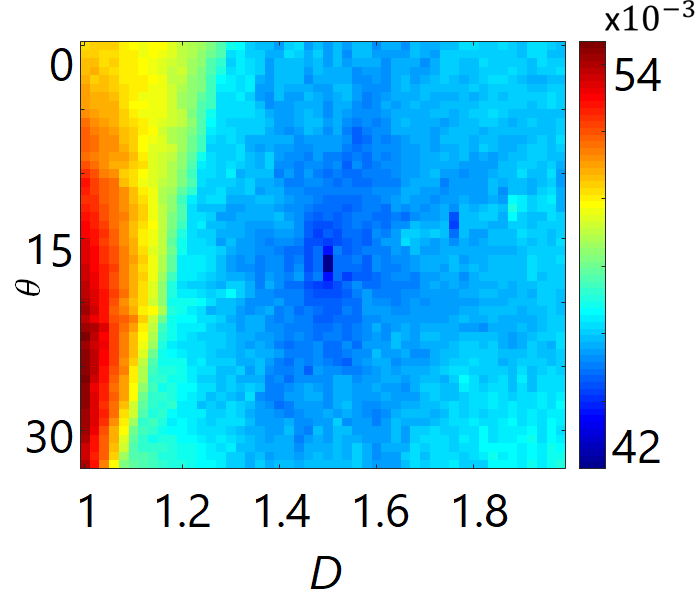} &
	\includegraphics[width=0.3\linewidth,trim={0cm 0cm 0cm 0cm},clip]{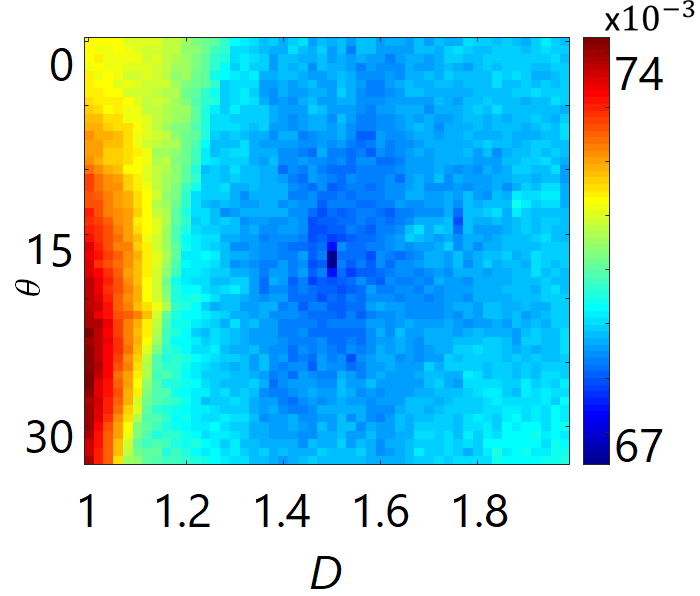} &
	\includegraphics[width=0.3\linewidth,trim={0cm 0cm 0cm 0cm},clip]{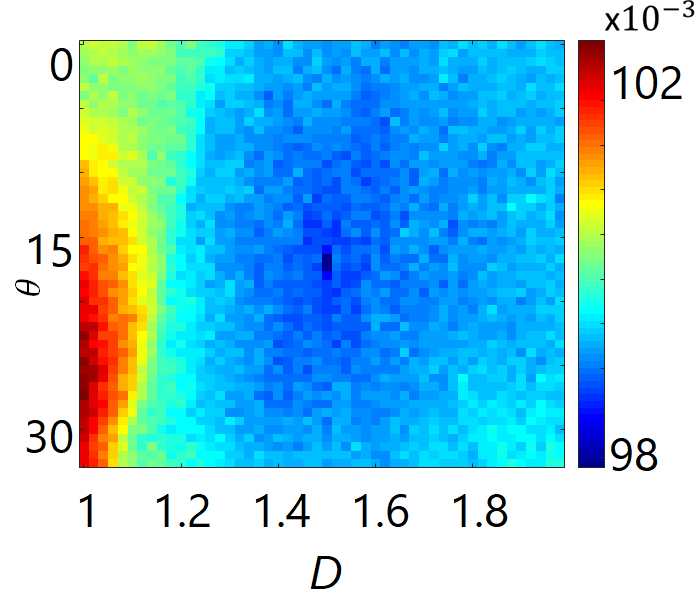} \\
	\includegraphics[width=0.3\linewidth,trim={0cm 0cm 0cm 0cm},clip]{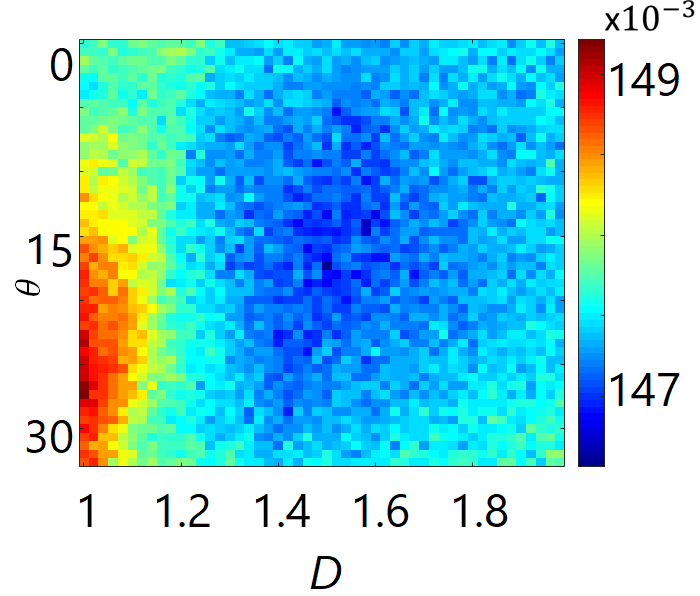} &
	\includegraphics[width=0.3\linewidth,trim={0cm 0cm 0cm 0cm},clip]{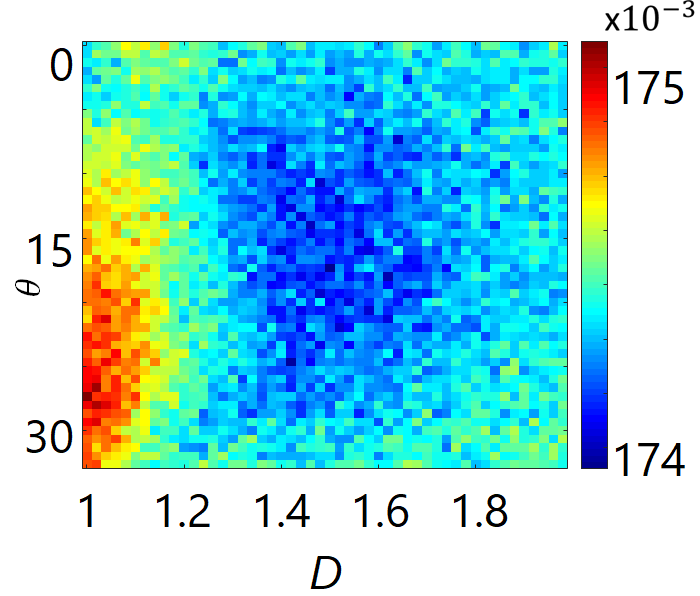} &
	\includegraphics[width=0.3\linewidth,trim={0cm 0cm 0cm 0cm},clip]{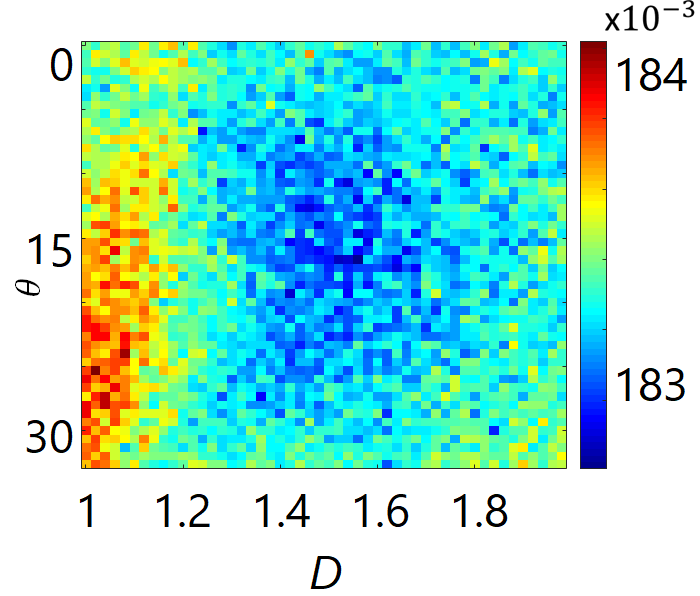} \\

    \multicolumn{3}{c}{Scene B} \\
	\includegraphics[width=0.3\linewidth,trim={0cm 0cm 0cm 0cm},clip]{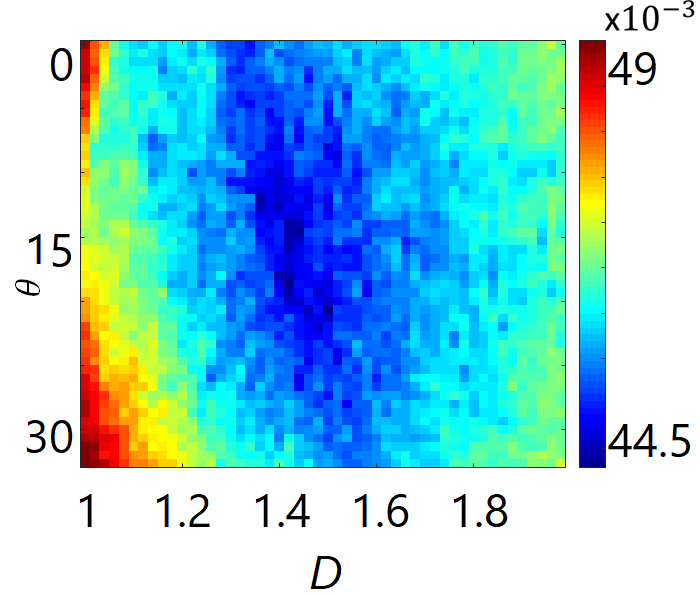} &
	\includegraphics[width=0.3\linewidth,trim={0cm 0cm 0cm 0cm},clip]{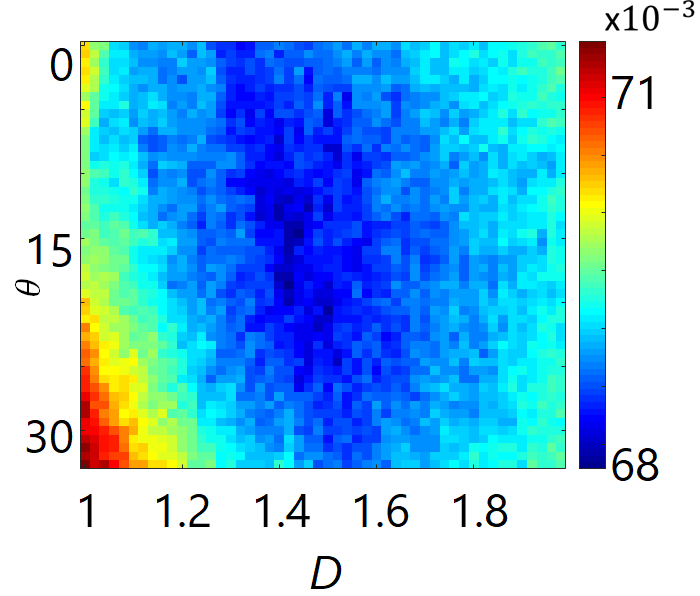} &
	\includegraphics[width=0.3\linewidth,trim={0cm 0cm 0cm 0cm},clip]{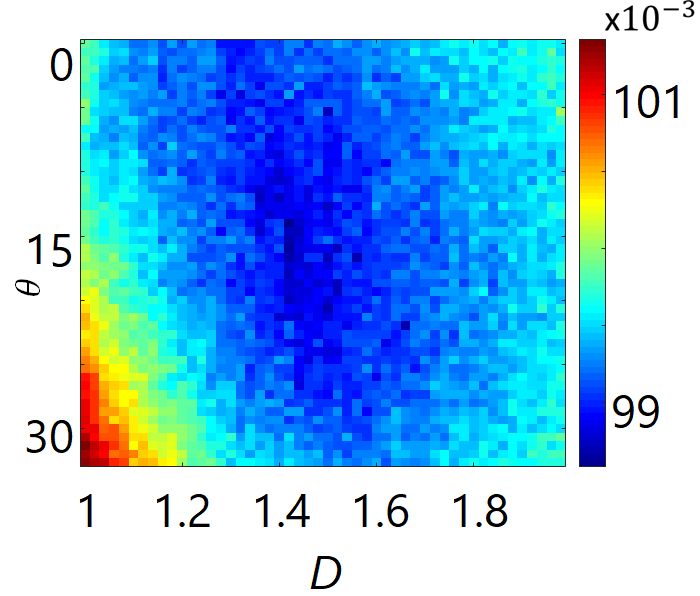} \\
	\includegraphics[width=0.3\linewidth,trim={0cm 0cm 0cm 0cm},clip]{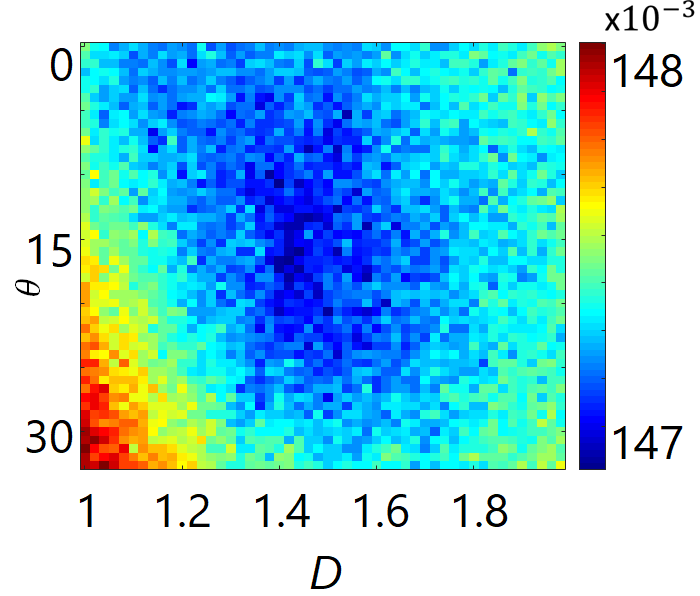} &
	\includegraphics[width=0.3\linewidth,trim={0cm 0cm 0cm 0cm},clip]{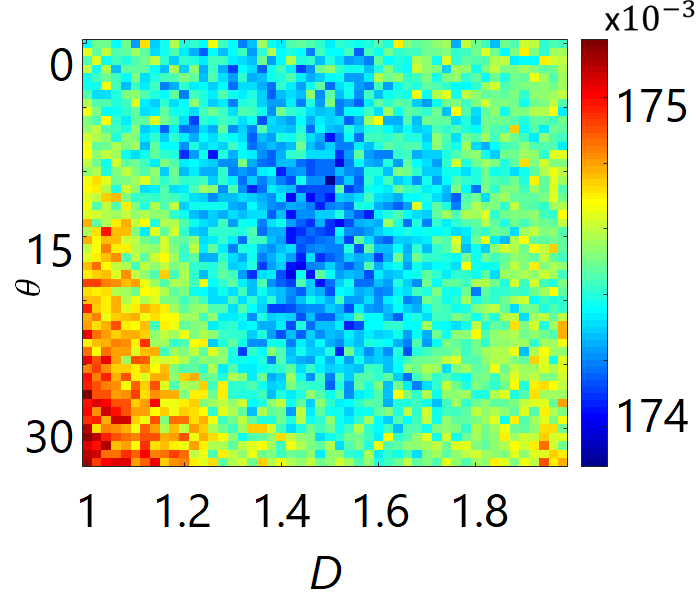} &
	\includegraphics[width=0.3\linewidth,trim={0cm 0cm 0cm 0cm},clip]{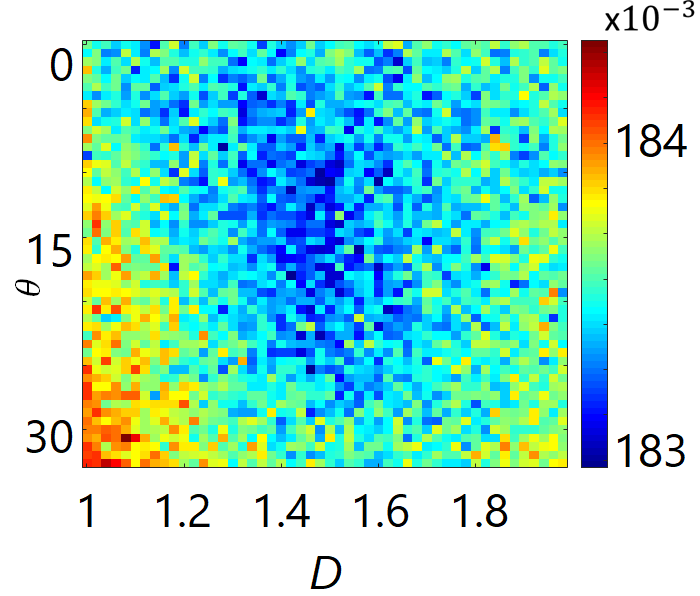} \\
	\end{tabular}
	\caption{Measure of the sparsity of the spectrum (RMSE of decimated spectrum, lower is better) depending on the image plane depth $D$ and orientation $\theta$ for scene A (top) and B (bottom), when adding Gaussian noise with $\sigma = 0.025, 0.05, 0.1, 0.25, 0.5, 0.75$ (from left to right, top to bottom). The minima observed in our previous experiments are still visible.}
	\label{fig:sparse_opt_vs_noise}
\end{figure}

In the next experiment, we keep testing the robustness of previous observations.
This time, we use the Lambertian texture with additive Gaussian noise, expressed as:
\begin{equation}
    \Lambda_n(x)=\Lambda(x) + \mathcal{N}(0,\sigma)
\end{equation}
\noindent where $\sigma$ represents the noise level.
The results for this experiment are shown in Figs.~\ref{fig:sparse_opt_vs_noise}.
We can again see that even in presence of noise, the spectrum is the most sparse when the image plane best fit the scene geometry, as the minima observed in the previous experiments are still visible, although for very high noise levels such as $\sigma=0.5$ or $\sigma=0.75$ the error gradient descending to the minimum is not as smooth.

\begin{figure}[t!]
	\centering
	\begin{tabular}{ccc}
	\multicolumn{3}{c}{Scene A} \\
	\includegraphics[width=0.3\linewidth,trim={0cm 0cm 0cm 0cm},clip]{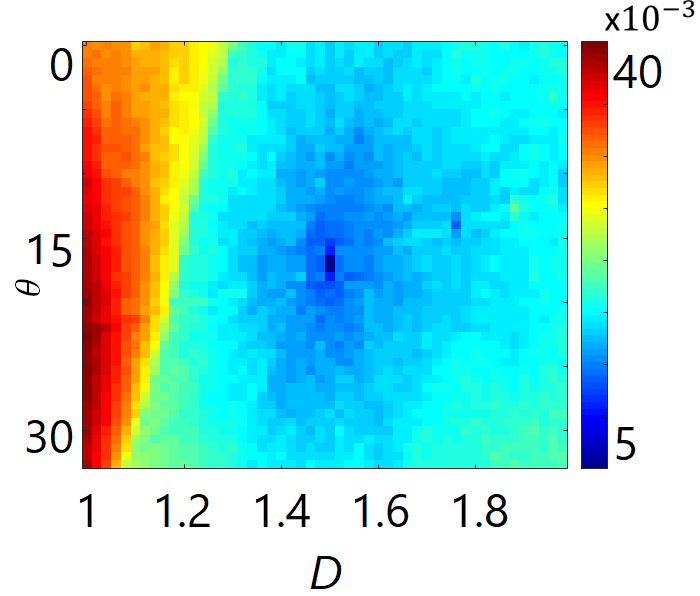} &
	\includegraphics[width=0.3\linewidth,trim={0cm 0cm 0cm 0cm},clip]{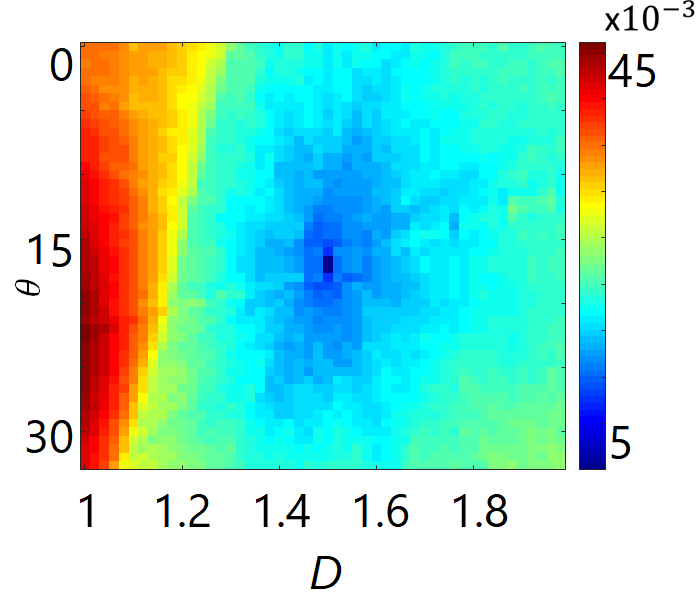} &
	\includegraphics[width=0.3\linewidth,trim={0cm 0cm 0cm 0cm},clip]{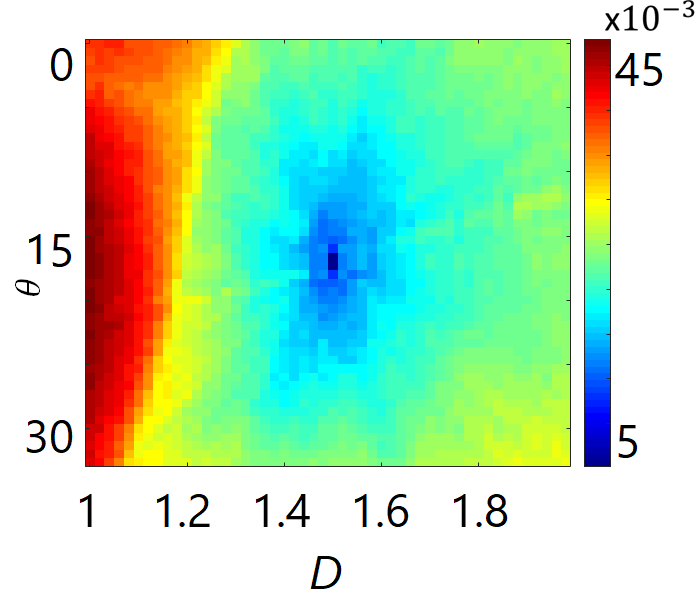} \\
	\includegraphics[width=0.3\linewidth,trim={0cm 0cm 0cm 0cm},clip]{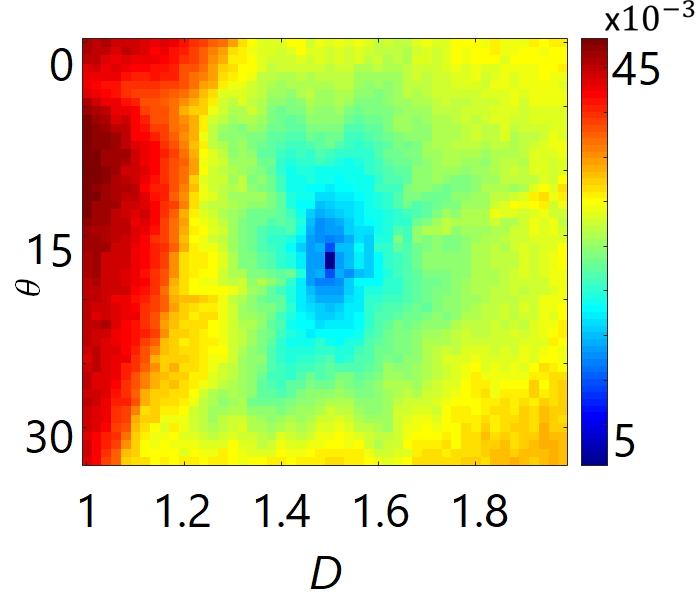} &
	\includegraphics[width=0.3\linewidth,trim={0cm 0cm 0cm 0cm},clip]{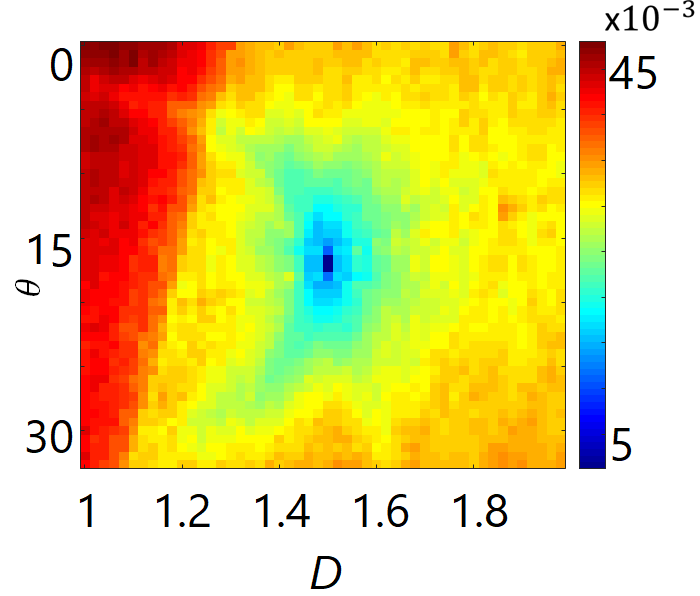} &
	\includegraphics[width=0.3\linewidth,trim={0cm 0cm 0cm 0cm},clip]{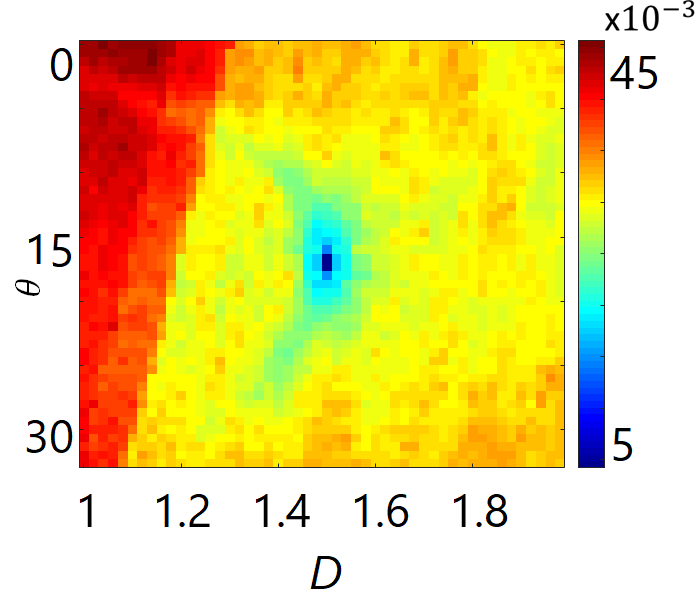} \\

    \multicolumn{3}{c}{Scene B} \\
	\includegraphics[width=0.3\linewidth,trim={0cm 0cm 0cm 0cm},clip]{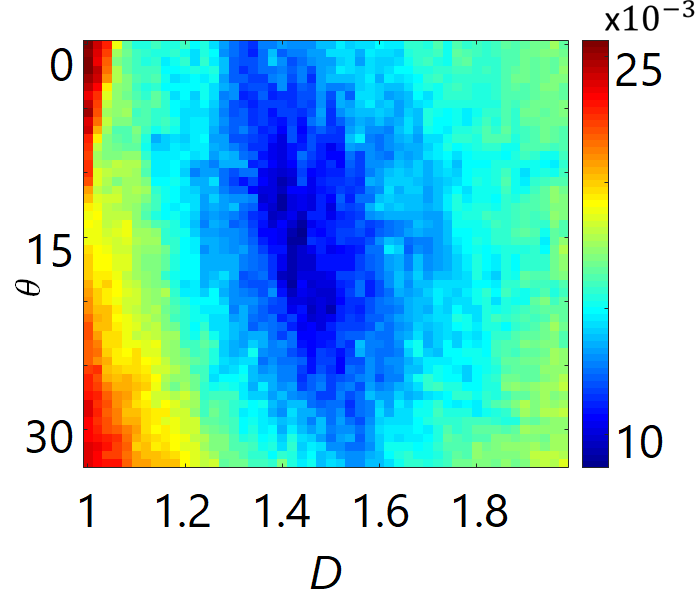} &
	\includegraphics[width=0.3\linewidth,trim={0cm 0cm 0cm 0cm},clip]{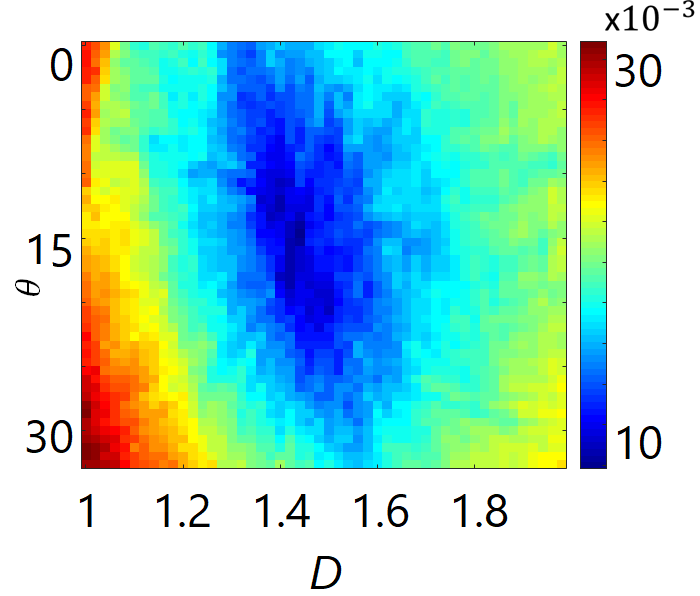} &
	\includegraphics[width=0.3\linewidth,trim={0cm 0cm 0cm 0cm},clip]{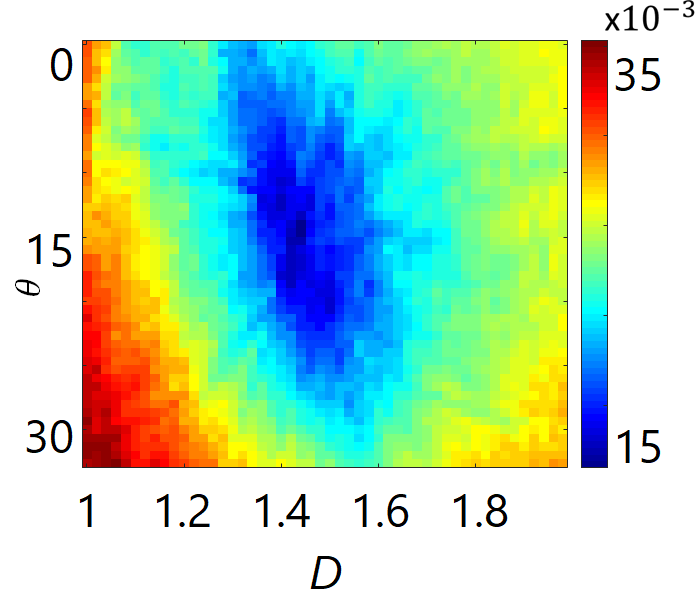} \\
	\includegraphics[width=0.3\linewidth,trim={0cm 0cm 0cm 0cm},clip]{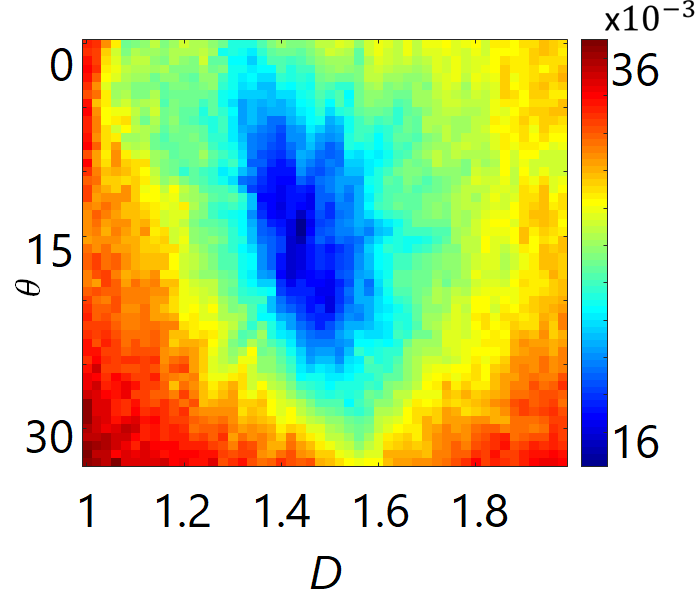} &
	\includegraphics[width=0.3\linewidth,trim={0cm 0cm 0cm 0cm},clip]{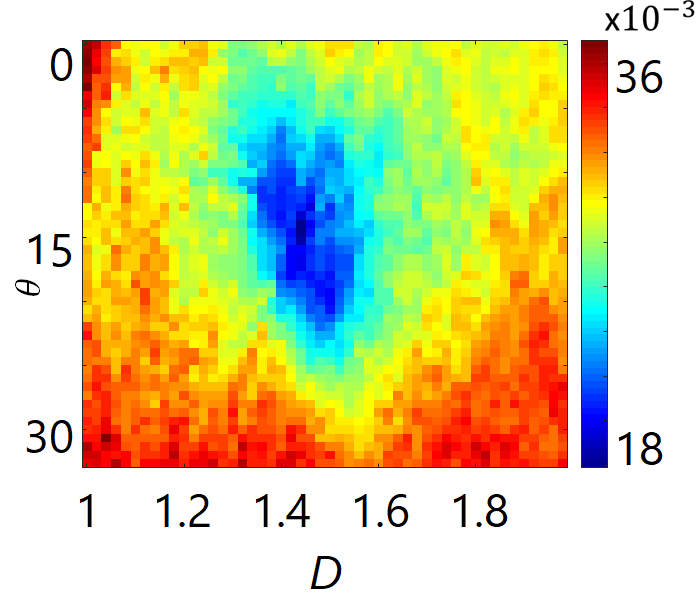} &
	\includegraphics[width=0.3\linewidth,trim={0cm 0cm 0cm 0cm},clip]{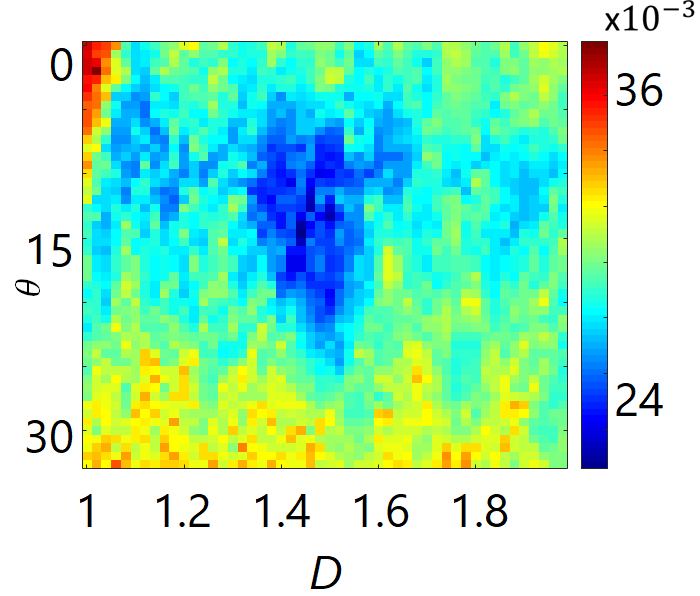} \\
	\end{tabular}
	\caption{Measure of the sparsity of the spectrum (RMSE of decimated spectrum, lower is better) depending on the image plane depth $D$ and orientation $\theta$ for scene A (top) and B (bottom), subsampling with factor $2, 4, 8, 16, 32, 64$ (from left to right, top to bottom). The minima observed in our previous experiments are still visible even for high subsampling factors.}
	\label{fig:sparse_opt_vs_subsampling}
\end{figure}

In this last experiment to study the sparsity of the light field depending on the image plane position and orientation, we use the Lambertian texture $\Lambda$, but test the impact of the density of the light field angular sampling.
For this purpose, we subsampled the original EPI containing 512 lines with factors 2, 4, 8, 16, 32, 64 along the angular dimension.
Results are shown in Figs.~\ref{fig:sparse_opt_vs_subsampling}, and further confirm the previous experiments, as the minima observed previously are again clearly visible, even when the light field is sparsely sampled.

\subsection{Evaluation of the re-parameterized light field reconstruction quality}

We now study how the re-parameterization of the light field can impact its reconstruction.
As in the previous experiment, we first subsampled the original EPI containing 512 lines with factors 2, 4, 8, 16, 32, 64 along the angular dimension.
We then reconstruct the light field to its original resolution using bilinear interpolation.
The quality of the reconstruction is measured using the Peak Signal to Noise Ratio (PSNR).
We use in this experiment scenes A and B with the Lambertian texture $\Lambda$.

\begin{figure}[t!]
	\centering
	\begin{tabular}{ccc}
	\multicolumn{3}{c}{Scene A} \\
	\includegraphics[width=0.3\linewidth,trim={0cm 0cm 0cm 0cm},clip]{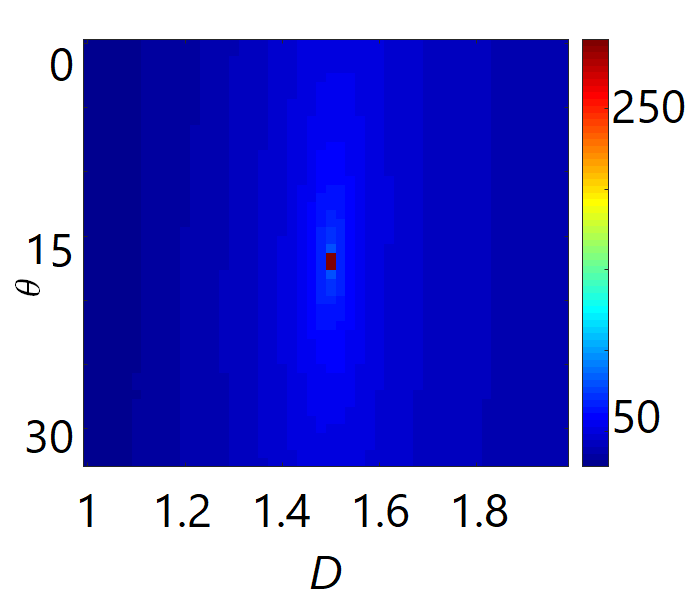} &
	\includegraphics[width=0.3\linewidth,trim={0cm 0cm 0cm 0cm},clip]{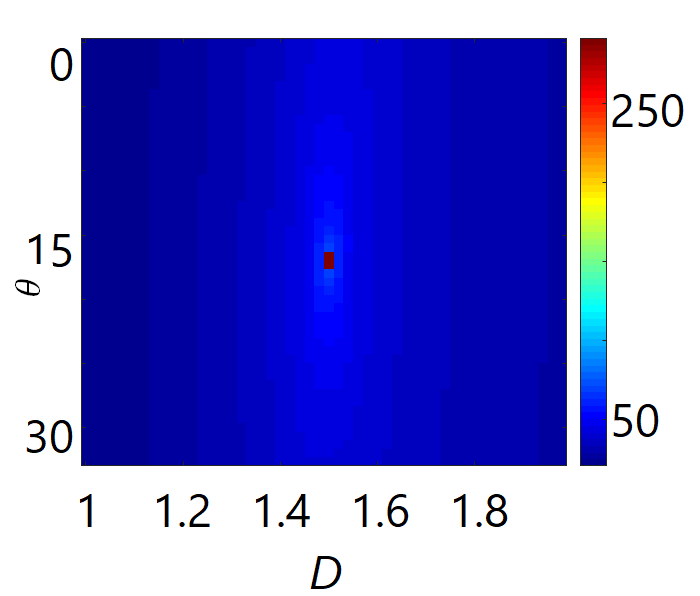} &
	\includegraphics[width=0.3\linewidth,trim={0cm 0cm 0cm 0cm},clip]{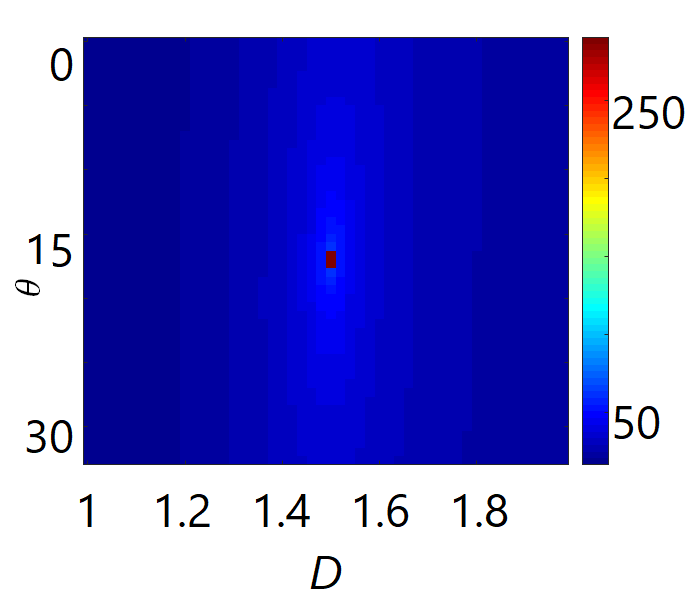} \\
	\includegraphics[width=0.3\linewidth,trim={0cm 0cm 0cm 0cm},clip]{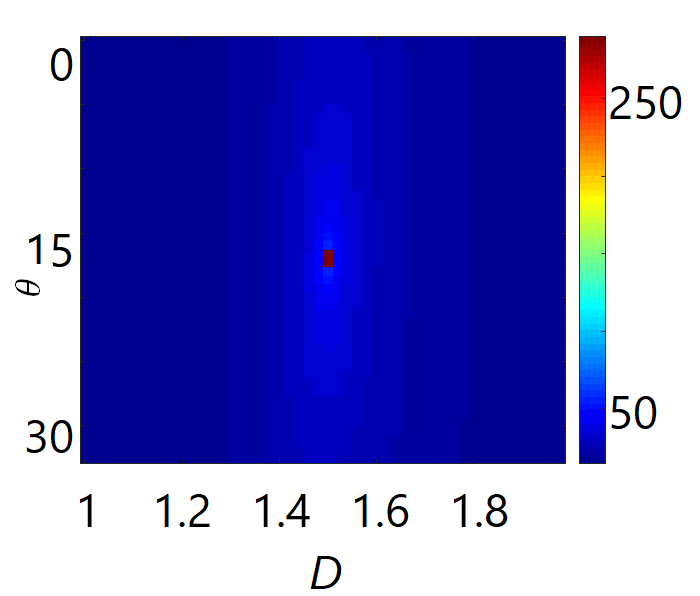} &
	\includegraphics[width=0.3\linewidth,trim={0cm 0cm 0cm 0cm},clip]{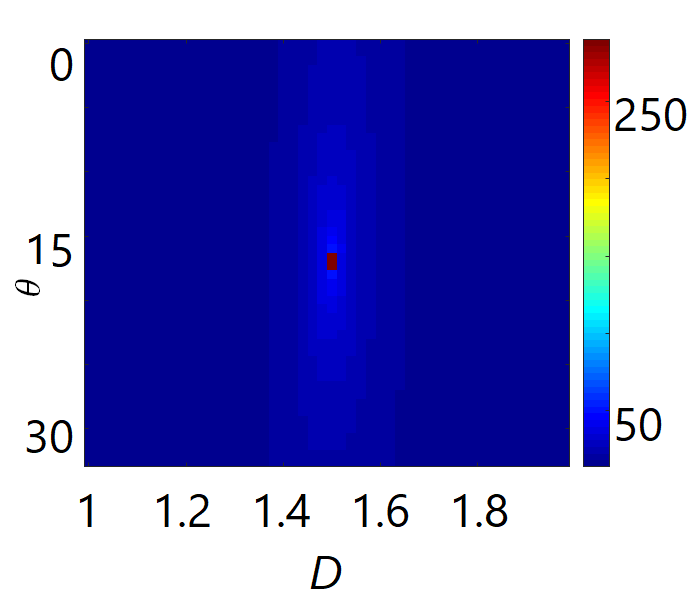} &
	\includegraphics[width=0.3\linewidth,trim={0cm 0cm 0cm 0cm},clip]{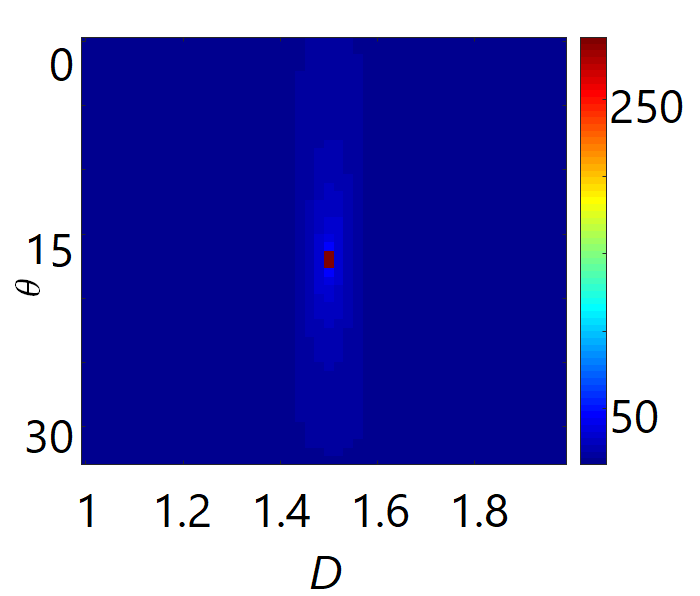} \\

    \multicolumn{3}{c}{Scene B} \\
	\includegraphics[width=0.3\linewidth,trim={0cm 0cm 0cm 0cm},clip]{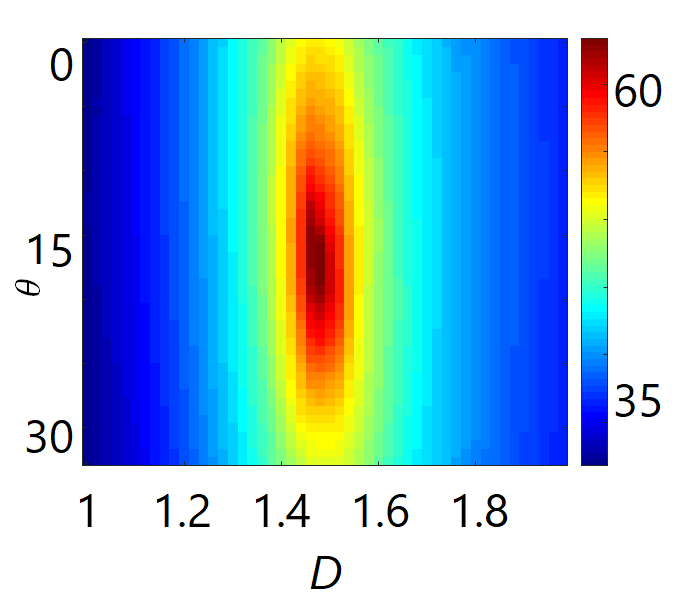} &
	\includegraphics[width=0.3\linewidth,trim={0cm 0cm 0cm 0cm},clip]{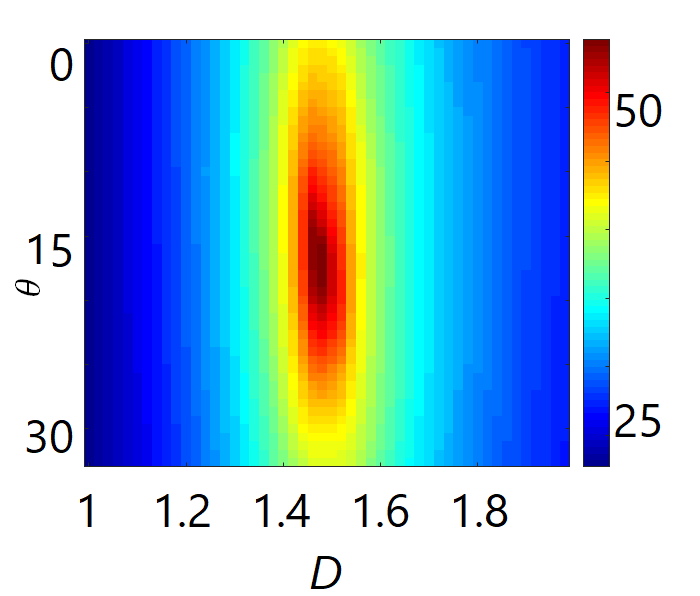} &
	\includegraphics[width=0.3\linewidth,trim={0cm 0cm 0cm 0cm},clip]{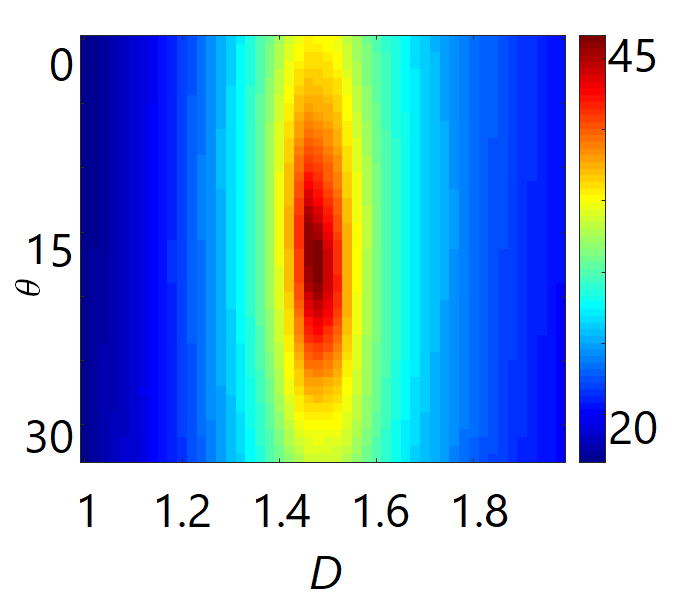} \\
	\includegraphics[width=0.3\linewidth,trim={0cm 0cm 0cm 0cm},clip]{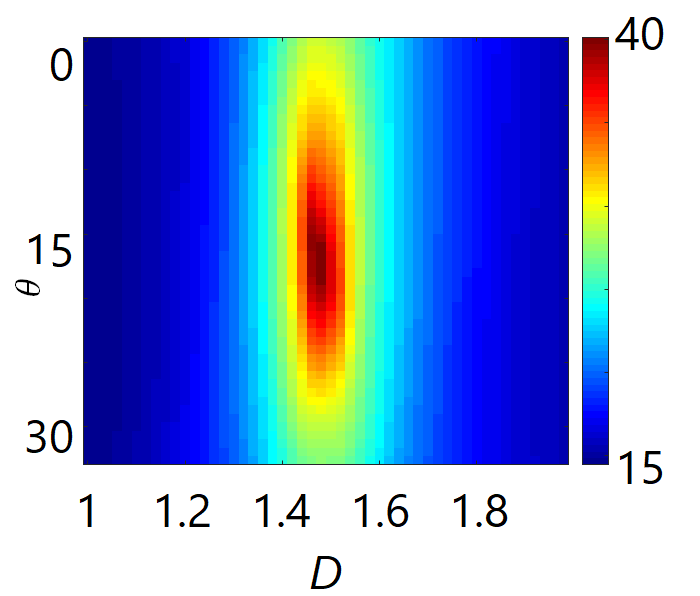} &
	\includegraphics[width=0.3\linewidth,trim={0cm 0cm 0cm 0cm},clip]{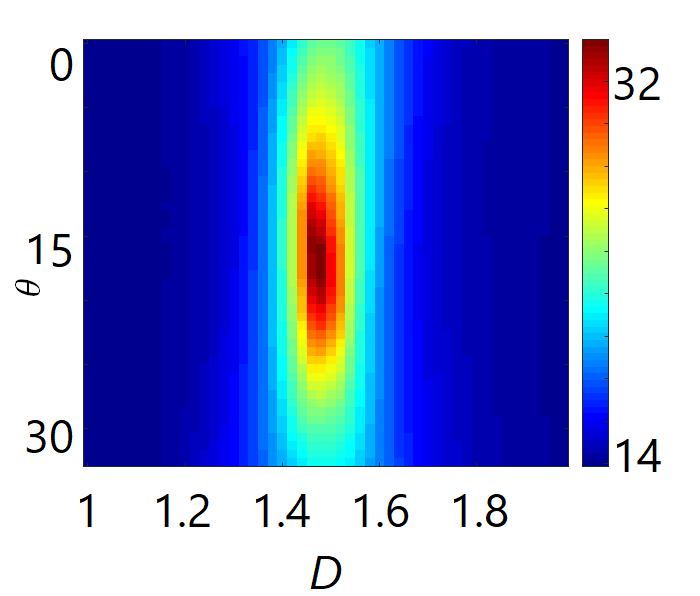} &
	\includegraphics[width=0.3\linewidth,trim={0cm 0cm 0cm 0cm},clip]{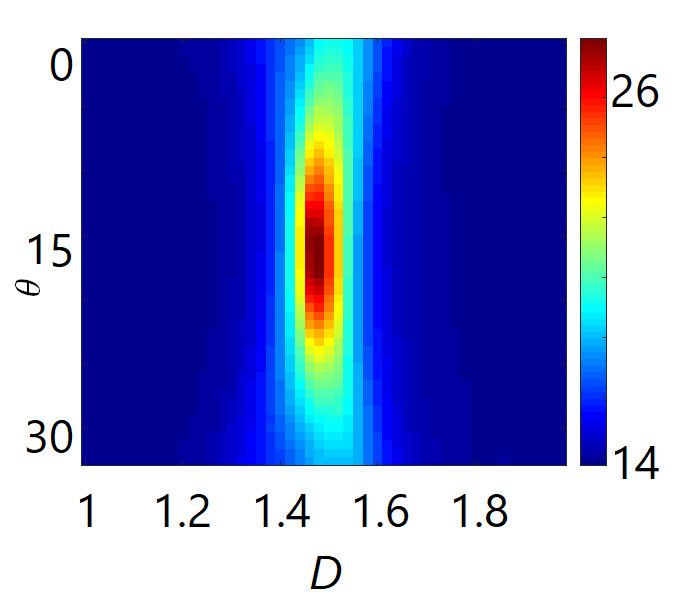} \\
	\end{tabular}
	\caption{Measure of the light field reconstruction quality (PSNR, higher is better) depending on the image plane depth $D$ and orientation $\theta$ for scene A (top) and B (bottom), subsampling with factor $2, 4, 8, 16, 32, 64$ (from left to right, top to bottom). For both scenes and all subsampling factors, the reconstruction quality is clearly best the image plane best fit the scene geometry.}

	\label{fig:rec_vs_subsampling}

\end{figure}

Results are shown in Figs.~\ref{fig:rec_vs_subsampling}.
As predicted in theory, for scene A which consists of a single tilted plane, when the image plane coincides with the scene object surface, almost perfect reconstruction is possible even when the light field is sparsely sampled.
For scene B, the reconstruction quality is also best when the image plane fits the scene geometry, which confirms that sampling guidelines are more advantageous than the parallel plane light field when using a tilted plane.

In our last experiment, we study again the light field reconstruction quality, but here we divide the scene geometry in multiple non-overlapping depth layers and evaluate the performance depending on the number of depth layers and subsampling factor.
We test from 1 to 16 depth layers. 
The original angular resolution is changed to 1024 lines, which are subsampled with factors $2, 3, 4, 6, 8, 12, 16, 24, 32, 48, 64, 96, 128, 194, 256$ and  $512$.
We compare the parallel plane and the tilted image plane parameterizations, such that the optimal plane depth and orientation are found for each depth layer.
We use in this experiment scene C with the Lambertian texture $\Lambda$.

\begin{figure}[t]
	\centering
	\begin{tabular}{ccc}
	\includegraphics[width=0.3\linewidth,trim={0cm 0cm 0cm 0cm},clip]{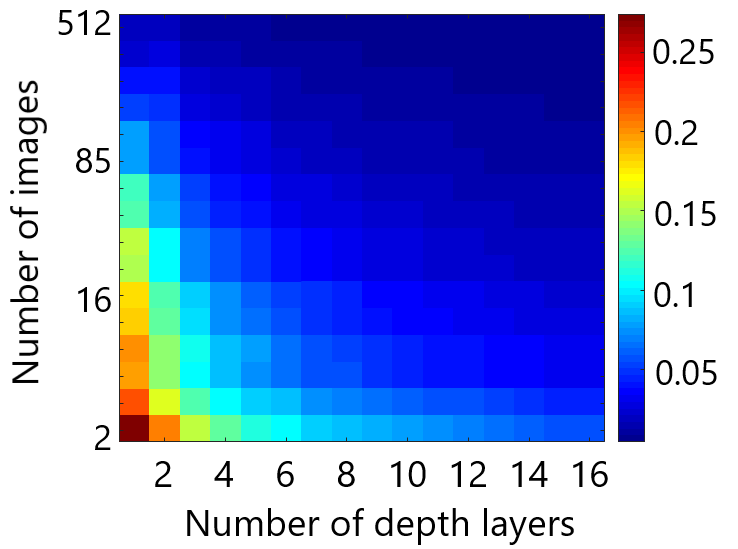} &
	\includegraphics[width=0.3\linewidth,trim={0cm 0cm 0cm 0cm},clip]{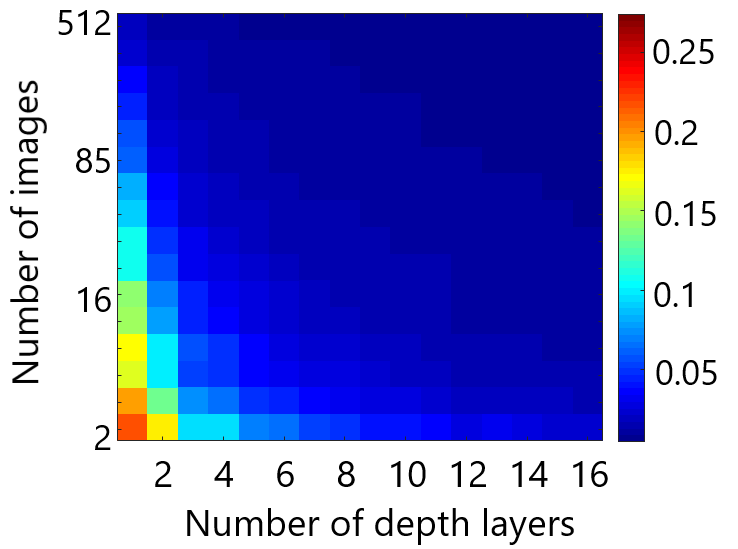} & 
	\includegraphics[width=0.3\linewidth,trim={0cm 0cm 0cm 0cm},clip]{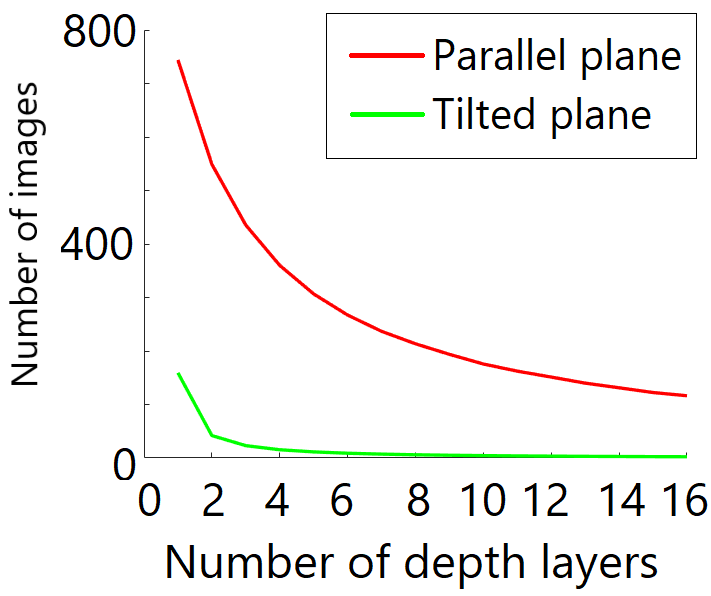} \\
	(a) Parallel image plane & (b) Tilted image plane & (c) Sampling curves\\
	\end{tabular}
	\caption{Measure of the light field reconstruction quality (RMSE, lower is better) depending on the number of depth layers and the number of images, for the parallel image plane (a) and tilted image plane (b). The reconstruction error is always lower for the tilted image plane. The corresponding theoretical sampling curves showing the minimum number of images required to achieve a dense light field sampling depending on the number of depth layers are plotted in (c).}
	\label{fig:multiplane}

\end{figure}

The reconstruction quality is measured using the RMSE between the original and reconstructed light fields for each test point, and results are shown in Fig.~\ref{fig:multiplane} (a) and (b) for the parallel and tilted image planes respectively.
We can see that the reconstruction error is lower for the tilted image plane, even when the light field is largely undersampled.
Furthermore, we show in  Fig.~\ref{fig:multiplane} (c) the theoretical sampling curves (as introduced by Chai et al. shown in Fig.~\ref{fig:EPISpectrumSampling}), obtained by first computing the maximum camera baselines using equations~\ref{eq:min_sampling_replica} and~\ref{eq:smax_bl_tilt}, and then deriving the corresponding number of images within the camera plane range $[-s_{max}, s_{max}]$.
These curves clearly show that fewer images are needed for all numbers of depth layers when using the appropriate tilted parameterization of the image plane compared to the parallel image plane.
This further demonstrates the advantage of the new sampling guidelines associated with the tilted image plane re-parameterization proposed in this paper.

\section{Conclusion}
\label{sec:conclusion}

We studied in this paper the spectral properties of light fields depending on the parameterization of its image plane.
In particular, we introduced an additional degree of freedom compared to existing two-parallel plane light fields by allowing to tilt the image plane.
We demonstrated that this can be useful by adapting the image plane to the scene geometry, which can lead to smaller spectrum support in the Fourier domain, and allows sampling the light field with larger camera baseline.

On the other hand, if the image plane is not aligned with the scene geometry, the light field spectrum support expands, which suggests that the spectrum sparsity could be used as a loss function in future work on scene geometry estimation.

Our simulations also showed that this can be useful when using multiple image planes, i.e. the geometry of the scene is known.
In practice, these findings could be used for optimizing multi-plane image (MPI) representations, which is an active research topic~\cite{Navarro2021}.

Another interesting finding arising from the theoretical spectral analysis of the tilted light field image plane is the emergence of the \textit{chirp} function, originally applied to frequency modulation in telecommunication applications.
The concept of chirp has been later extended to a chirplet transform~\cite{Mann1995}.
While the chirplet transform has also been mainly applied to radar signal processing, it is also of interest to obtain sparse representations of perspective images of periodic structures.
In other words, an image of a periodic texture pasted on a tilted plane can be efficiently represented using the chirplet transform.
In previous analysis of the two-parallel plane light field spectrum, the counterpart of the chirp function was simply a complex phase shift function, which effect is to shear the light field spectrum support.
The observation of this shearing effect has led to the development of a compact light field representation based on the shearlet transform~\cite{vagharshakyan2017light}.
Arguably, the emergence of the chirp function in our analysis could thus lead to a novel, more compact, light field representation based on the chirplet transform.

Finally, as this paper explored the theoretical spectral analysis of re- parameterized light fields for the first time, we have relied on a common but simplified model of the scene, e.g. not considering occlusions. As it was shown by Zhang and Chen in~\cite{zhang2003spectral} that the impact of occlusions is to widen the spectrum support, we believe that our new sampling guideline could still be more benefitial than the two-parallel plane parameterization when considering occlusions. In future work, this could be demonstrated by extending our analysis to explicitly handle occlusions as in~\cite{zhang2003spectral}, where the scene is modeled as a set of planar objects parallel to the camera plane. Furthermore, our tilted image plane approach would be conducive to synthesize the model of Zhang and Chen with the work of Gillam et al.~\cite{gilliam2013spectrum}, where the scene is modeled as a set of tilted planes, but occlusions are not considered.

\clearpage


%

\appendix
\section{Development for the two-parallel plane parameterization (section \ref{sec:TPP})}
\label{sec:TPP_app}

\subsection{Geometric mapping}

\begin{figure}[t]
	\centering
	\includegraphics[width=\linewidth,trim={0cm 1.1cm 0cm 0cm},clip]{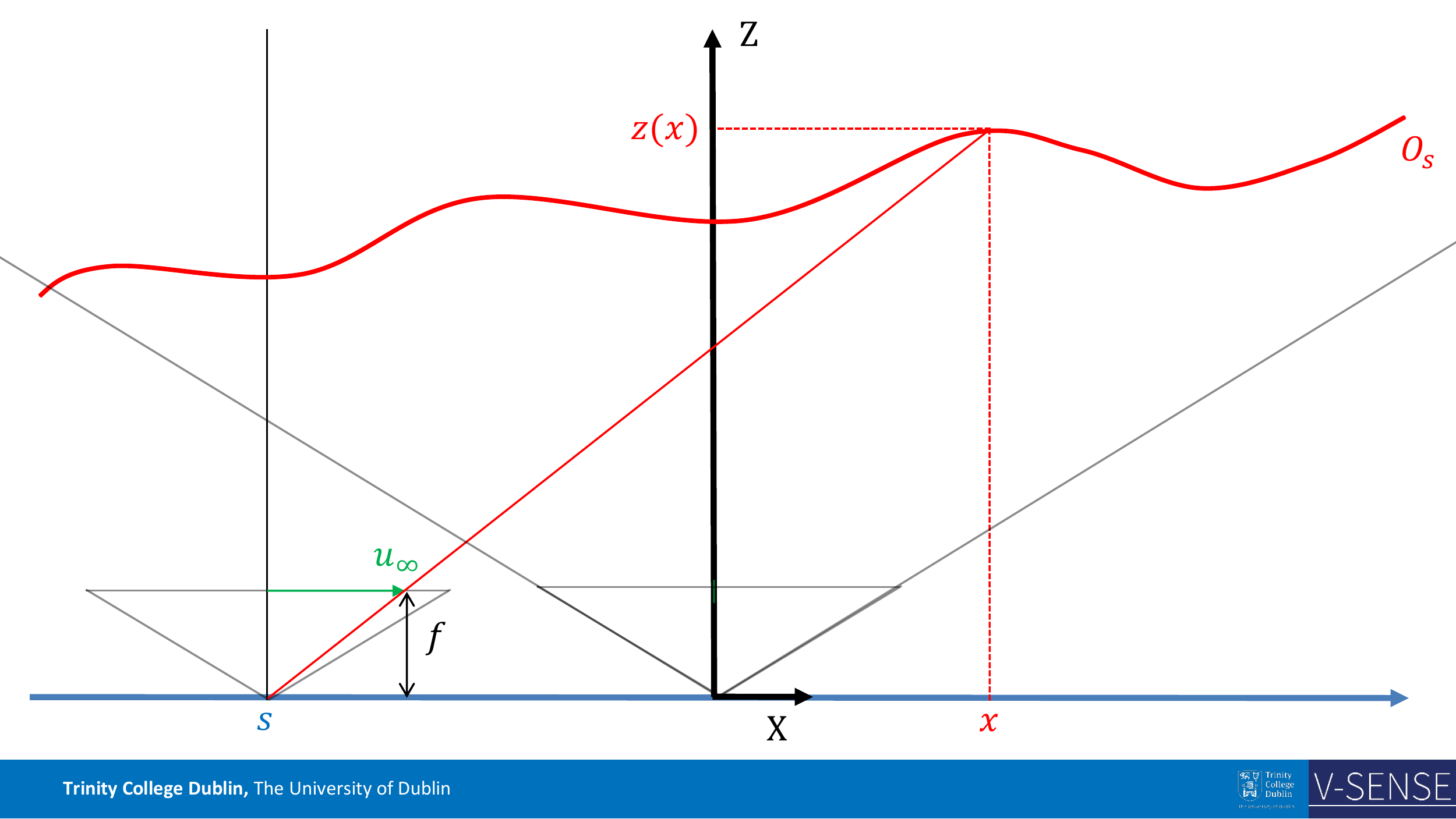}
	\caption{Two-Parallel Plane parameterization for a 2D slice of the light field when the global image plane is at infinity. }
	\label{fig:uinf_similar_triangle}
\end{figure}

The relation defined in equation~\ref{eq:lf_geo_u_inf} comes from the observation of the similar triangles visible in Fig.~\ref{fig:uinf_similar_triangle}, which yields:

\begin{equation}
    \frac{x - s}{u_{\infty}} = \frac{z(x)}{f}
\end{equation}

\noindent By substituting in equation~\ref{eq:lf_geo_u_inf} the expression of $u_{\infty}$, defined in equation~\ref{eq:u_inf}, we obtain:

\begin{equation}
    \label{eq:app:lf_geo_dev}
    \frac{z(x)u}{f} = \frac{z(x)s}{D} + x - s
    = x + s (\frac{z(x)}{D} - 1)
\end{equation}

\noindent By multiplying by $\frac{f}{z(x)}$ on both sides of equation~\ref{eq:app:lf_geo_dev} to isolate $u$ on the left-hand side, we obtain the geometric mapping defined in equation~\ref{eq:lf_geo}.

\subsection{No self-occlusion condition}

The no self-occlusion condition is derived in~\cite{do2011bandwidth} to ensure that each light ray intersects the object surface only once, i.e. the geometric mapping derived above is a one-to-one mapping.
Intuitively, this means that the slope of the object surface is bounded by the maximum viewing angle.
This is expressed mathematically, by observing that this is equivalent to requiring that $s$ in equation~\ref{eq:lf_geo_u_inf} is a monotonic function of $x$.
We can rewrite equation~\ref{eq:lf_geo_u_inf} to express $s$ depending on $x$ as:

\begin{equation*}
    s = x - \frac{z(x)}{f}u_\infty
\end{equation*}

By taking the derivative of this expression, we get:

\begin{equation*}
    \frac{ds}{dx} = 1 - \frac{z'(x)}{f}u_\infty
\end{equation*}

The monotonic requirement can then be expressed as:

\begin{equation*}
    \frac{ds}{dx} \geq 0
\end{equation*}

\noindent which gives:

\begin{equation}
    \label{eq:no_self_occ_uinf}
    \lvert z'(x) \rvert \leq \frac{f}{\lvert u_\infty \rvert}
\end{equation}

In this paper, we consider that the local image plane is limited by $\lvert u \rvert < u_{max}$, and the camera plane is limited by $\lvert s \rvert < s_{max}$.
Thus, using equation~\ref{eq:u_inf}, the maximum viewing angle can be expressed as:

\begin{equation*}
    \max(\lvert u_\infty \rvert) = u_{max} + \frac{s_{max}f}{D}
\end{equation*}

Finally, by substituting this expression in equation~\ref{eq:no_self_occ_uinf}, we get the no-self occlusion condition of equation~\ref{eq:no_self_occ}.

\subsection{Light field spectrum}

We show here the derivations leading to the expression of equation~\ref{eq:P_fourier_tr}.
First, let's express the derivative of $u$ with respect to $x$ by differentiating equation~\ref{eq:lf_geo}:
\begin{equation}
\label{eq:du_dx}
\begin{split}
\frac{du}{dx} = \frac{f (z(x) - z'(x) x)}{z(x)^2} + \frac{s f z'(x)}{z(x)^2} 
= \frac{f}{z(x)^2}(z(x) + z'(x) (s-x))
\end{split}
\end{equation}

Starting from the definition of $P$,
\begin{equation*}
    P(\omega_s, \omega_u) \triangleq \mathcal{F}_{s,u}\{p(s,u)\} 
    = \int_{-\infty}^{\infty} \int_{-\infty}^{\infty} p(s,u) e^{-j(\omega_s s + \omega_u u)} ds du
\end{equation*}

\noindent and using the change of variable based on equations~\ref{eq:lf_geo} and \ref{eq:du_dx}, we have:

\begin{multline}
\label{eq:P_fourier_tr_dev}    
    P(\omega_s, \omega_u) 
    = 
    \int_{-\infty}^{\infty} \int_{-\infty}^{\infty} l(x,s) e^{-j\left(\omega_s s + \omega_u \left(\frac{xf}{z(x)} + s f (\frac{1}{D} - \frac{1}{z(x)})\right) \right)} \\ \frac{f}{z(x)^2}(z(x) + z'(x) (s-x)) ds dx \\
    =
    \int_{-\infty}^{\infty}  \frac{f}{z(x)^2} e^{-j\left(\omega_u \frac{xf}{z(x)}\right)} \bigl( \int_{-\infty}^{\infty} \left(z(x) + z'(x) (s-x)\right) l(x,s) \\ e^{-j\omega_s s} e^{-j \omega_u f (\frac{1}{D} - \frac{1}{z(x)}) s} ds \bigr) dx \\
    =
    \int_{-\infty}^{\infty}  \frac{f}{z(x)^2} e^{-j\left(\omega_u \frac{xf}{z(x)}\right)} \bigl( \int_{-\infty}^{\infty} h(x,s) g(x,s) e^{-j\omega_s s} ds \bigr) dx 
\end{multline}

\noindent where $h(x,s) \triangleq (z(x) + z'(x) (s-x)) l(x,s)$, and $g(x,s) \triangleq  e^{-j \omega_u f (\frac{1}{D} - \frac{1}{z(x)}) s}$.
The Fourier transform of $h$ and $g$ over $s$ are defined as:

\begin{equation*}
    H(x, \omega_s) \triangleq \mathcal{F}_{s}\{h(x,s)\} 
    = \int_{-\infty}^{\infty} h(x,s) e^{-j\omega_s s} ds \\
\end{equation*}
\begin{equation*}
    G(x, \omega_s) \triangleq \mathcal{F}_{s}\{g(x,s)\} 
    = \int_{-\infty}^{\infty} g(x,s) e^{-j\omega_s s} ds \\
\end{equation*}

\noindent and from the properties of the Fourier transform, we have:

\begin{equation*}
    \int_{-\infty}^{\infty} h(x,s) g(x,s) e^{-j\omega_s s} ds 
    = H(x, \omega_s) * G(x, \omega_s)
\end{equation*}

\noindent where $*$ is the convolution over $\omega_s$. Furthermore, we have an analytic formulation for $G$:

\begin{equation*}
    G(x, \omega_s) = 2\pi\delta(\omega_s+\omega_u f (\frac{1}{D} - \frac{1}{z(x)}))
\end{equation*}

\noindent where $\delta$ is the Dirac delta function.
And from the properties of the convolution we can see that:

\begin{equation*}
    \int_{-\infty}^{\infty} h(x,s) g(x,s) e^{-j\omega_s s} ds = H\left(x, \omega_s + \omega_u f (\frac{1}{D} - \frac{1}{z(x)})\right)
 \end{equation*}
 
 By substituting in equation~\ref{eq:P_fourier_tr_dev}, we obtain the final expression of equation~\ref{eq:P_fourier_tr}.
 
 \section{Development for the tilted image plane parameterization (section \ref{sec:TIP})}
\label{sec:TIP_app}

\subsection{Geometric mapping}

\begin{figure}[t]
	\centering
	\includegraphics[width=\linewidth,trim={0cm 1.1cm 0cm 0cm},clip]{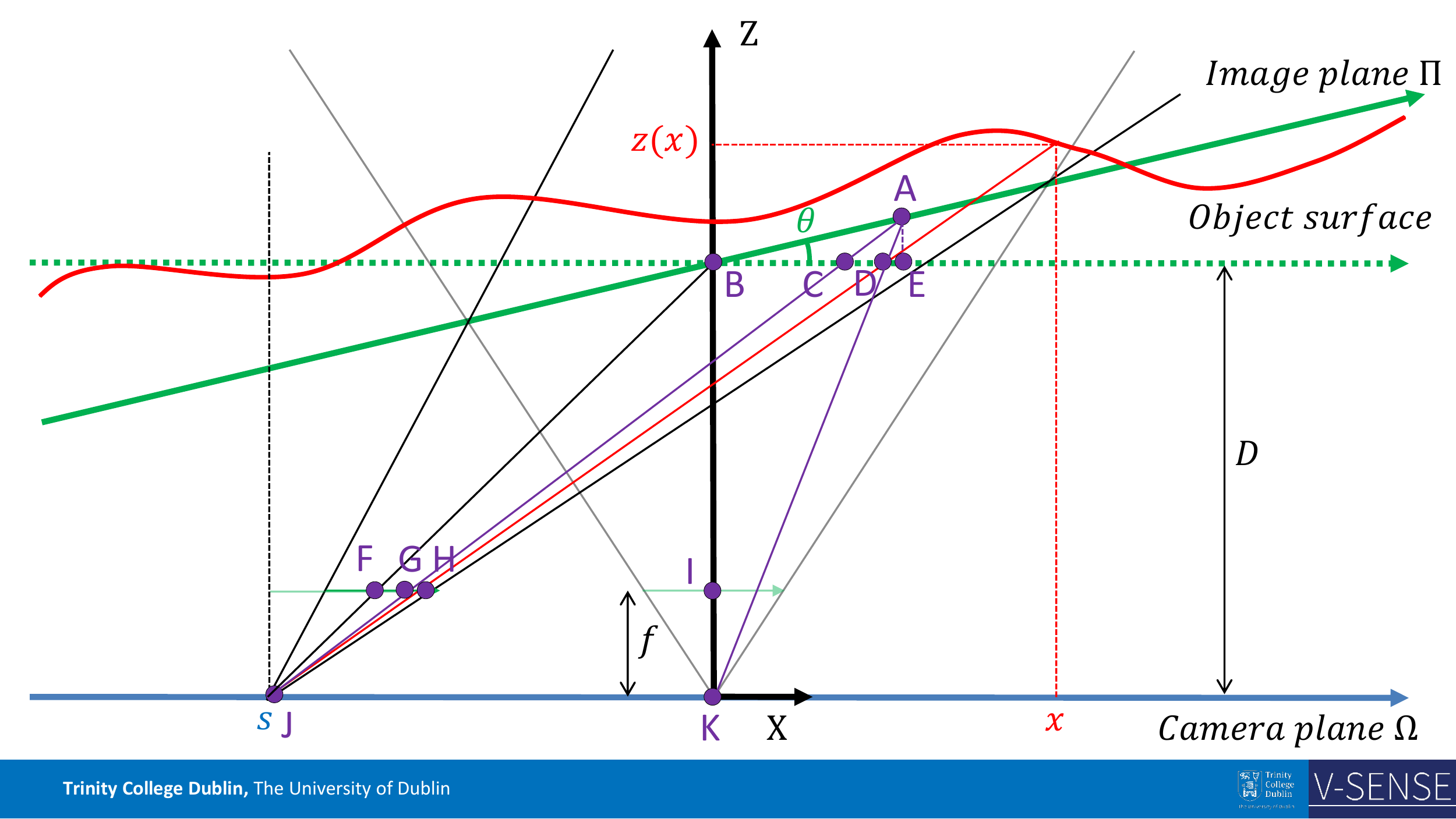}
	\caption{Tilted Image Plane parameterization for a 2D slice  of the light field.}
	\label{fig:TIP_annotated}
\end{figure}

The relation defined in equation~\ref{eq:u_inf_tilt} can be obtained by combining a few simple geometric observations.
For simplicity, we show in Fig.~\ref{fig:TIP_annotated} a version of Fig.~\ref{fig:TIP} with annotated points from A to K.
We can then observe that the following pairs of triangles are similar: ADE and KDB, ACD and AJK, JFH and JBD, and JGH and JCD.
This yields, respectively:

\begin{equation}
    \label{eq:app:DE}
    \frac{\text{DE}}{\text{BD}} = \frac{\text{EA}}{\text{KB}}
\end{equation}

\begin{equation}
    \label{eq:app:CD}
    \frac{\text{CD}}{\text{JK}} = \frac{\text{EA}}{\text{KB} + \text{EA}}
\end{equation}

\begin{equation}
    \label{eq:app:uP}
    \frac{\text{FH}}{\text{BD}} = \frac{\text{KI}}{\text{KB}}
\end{equation}

\begin{equation}
    \label{eq:app:uPu}
    \frac{\text{GH}}{\text{CD}} = \frac{\text{KI}}{\text{KB}}
\end{equation}

\noindent Furthermore, we can see that:

\begin{equation}
    \label{eq:app:AE}
    \text{EA} = \text{BE}\tan(\theta) = (\text{BD + DE})\tan(\theta)
\end{equation}

\noindent By combining equations \ref{eq:app:AE} and \ref{eq:app:DE}, we get:

\begin{equation*}
    \text{EA} = (\text{BD} + \frac{\text{BD}}{\text{KB}}\text{EA})\tan(\theta)
\end{equation*}

\begin{equation*}
    \text{EA}(1-\frac{\text{BD}}{\text{KB}}\text{EA}) = 
    \text{BD}\tan(\theta)
\end{equation*}

\begin{equation*}
    \text{EA} = 
    \frac{\text{KB}\cdot\text{BD}\tan(\theta)}{\text{KB}-\text{BD}\tan(\theta)}
\end{equation*}

\noindent By substituting in equation~\ref{eq:app:CD}, we then get:

\begin{equation*}
    \frac{\text{CD}}{\text{JK}} = 
    \frac{\text{KB}\cdot\text{BD}\tan(\theta)}{\text{KB}\cdot\text{BD}\tan(\theta) + \text{KB}(\text{KB}-\text{BD}\tan(\theta))}
\end{equation*}

\begin{equation*}
    \text{CD} = \text{JK}  
    \frac{\tan(\theta)}{\text{KB}}\text{BD}
\end{equation*}

\noindent By then substituting equations~\ref{eq:app:uPu} and~\ref{eq:app:uP} in the result above, we obtain:

\begin{equation*}
    \text{GH} \frac{\text{KB}}{\text{KI}} = \text{JK}  
    \frac{\tan(\theta)}{\text{KB}}\text{FH}\frac{\text{KB}}{\text{KI}}
\end{equation*}

The annotated points above are related to the rest of this paper notations as follows: $\text{KB} = D$, $\text{KI} = f$, $\text{JK} = -s$, $\text{FG} = u$.
In addition, we define $u_P$ as the intermediary coordinate on the camera sensor obtained from the two-parallel plane parameterization, and we have $\text{FH} = u_P$, and $\text{GH} = u_P - u$.
Thus, the equation above can be re-written as:

\begin{equation*}
    (u_P - u) \frac{D}{f} = -s  
    \frac{\tan(\theta)}{D}u_P\frac{D}{f}
\end{equation*}

\noindent which reduces to:

\begin{equation}
    \label{eq:app:u}
    u  = (1 + \frac{s\tan(\theta)}{D})u_P
\end{equation}

By definition of $u_P$, we can express $u_\infty$ depending on $u_P$ as in equation~\ref{eq:u_inf}:

\begin{equation*}
    u_\infty  = u_P - \frac{sf}{D}
\end{equation*}

\noindent By substituting into equation~\ref{eq:app:u}, we can finally obtain the relation defined in equation~\ref{eq:u_inf_tilt}.

\subsection{Light field spectrum}

We show here the derivations leading to the expression of equation~\ref{eq:P_fourier_tr_tilt}.
First, let's express the derivative of $u$ with respect to $x$ by differentiating equation~\ref{eq:lf_geo_tilt}:

\begin{equation}
\label{eq:du_dx_tilt}
\begin{split}
\frac{du}{dx} = (1 + \frac{s\tan(\theta)}{D})\frac{f}{z(x)^2}(z(x) + z'(x) (s-x))
\end{split}
\end{equation}

Starting from the definition of $P$,
\begin{equation*}
    P(\omega_s, \omega_u) \triangleq \mathcal{F}_{s,u}\{p(s,u)\} 
    = \int_{-\infty}^{\infty} \int_{-\infty}^{\infty} p(s,u) e^{-j(\omega_s s + \omega_u u)} ds du
\end{equation*}

\noindent and using the change of variable based on equations~\ref{eq:lf_geo_tilt} and \ref{eq:du_dx_tilt}, we have:

\begin{multline}
\label{eq:P_fourier_tr_dev_tilt}    
    P(\omega_s, \omega_u) 
    = 
    \int_{-\infty}^{\infty} \int_{-\infty}^{\infty} l(x,s) e^{-j\left(\omega_s s + \omega_u \left(1 + \frac{s\tan(\theta)}{D}  \right)\left(\frac{xf}{z(x)} + s f \left(\frac{1}{D} - \frac{1}{z(x)}\right)\right) \right)} \\  (1 + \frac{s\tan(\theta)}{D})\frac{f}{z(x)^2}(z(x) + z'(x) (s-x)) ds dx \\
    =
    \int_{-\infty}^{\infty}  \frac{f}{z(x)^2} e^{-j\omega_u \frac{xf}{z(x)}} \bigl(  \int_{-\infty}^{\infty}  (1 + \frac{s\tan(\theta)}{D})\left(z(x) + z'(x) (s-x)\right) l(x,s) \\ e^{-j\omega_u f s\left(\left(1 + \frac{s\tan(\theta)}{D}  \right)\left(\frac{1}{D} - \frac{1}{z(x)}\right) + \frac{\tan(\theta)}{D}\frac{x}{z(x)}\right) } e^{-j\omega_s s} ds \bigr) dx \\
    =
    \int_{-\infty}^{\infty}  \frac{f}{z(x)^2} e^{-j\omega_u \frac{xf}{z(x)}} \bigl( \int_{-\infty}^{\infty} h_\theta(x,s)c(x,s) e^{-j\omega_s s} ds \bigr) dx \\
\end{multline}

\noindent where:

\begin{equation*}
    h_\theta(x,s) \triangleq (1 + \frac{s\tan(\theta)}{D})(z(x) + z'(x) (s-x)) l(x,s)    
\end{equation*}
\begin{equation*}
    c(x,s) \triangleq e^{j\left((\omega_u f (\frac{1}{z(x)} - \frac{1}{D} - \frac{\tan(\theta)x}{D z(x)})) s + \omega_u f \frac{\tan(\theta)}{D} (\frac{1}{z(x)} - \frac{1}{D}) s^2 \right)}    
\end{equation*}



\clearpage

\section*{References}
\bibliography{refs}

\end{document}